\documentclass{article}

\usepackage{times}
\usepackage{caption}
\usepackage{subcaption}
\usepackage{epsfig}
\usepackage{graphicx}
\usepackage{amsmath}
\usepackage{amssymb}
\usepackage{array}
\usepackage{colortbl}
\usepackage{booktabs}
\usepackage{bbm}
\usepackage{multirow}
\usepackage{multicol}
\usepackage{makecell}
\usepackage{xcolor}
\usepackage{xspace}
\usepackage{float}
\usepackage{color}
\usepackage{algorithm}
\usepackage{listings}
\usepackage{amsthm}

\usepackage[pagebackref=true,breaklinks=true,colorlinks,bookmarks=false]{hyperref}

\newcommand{\tran}{{^{\mkern-1.5mu\mathsf{T}}}}
\newcommand{\myparagraph}[1]{{\vspace{.5em} \noindent \bf #1}}
\newcommand{\eg}{\textit{e.g.}}
\newcommand{\ie}{\textit{i.e.}}

\usepackage[accepted]{icml2021}

\begin{document}

\twocolumn[
\icmltitle{What Makes for End-to-End Object Detection?}




\begin{icmlauthorlist}
\icmlauthor{Peize Sun}{hku}
\icmlauthor{Yi Jiang}{bytedance}
\icmlauthor{Enze Xie}{hku}
\icmlauthor{Wenqi Shao}{cuhk}
\icmlauthor{Zehuan Yuan}{bytedance}
\icmlauthor{Changhu Wang}{bytedance}
\icmlauthor{Ping Luo}{hku}
\end{icmlauthorlist}

\icmlaffiliation{hku}{Department of Computer Science, The University of Hong Kong}
\icmlaffiliation{bytedance}{AI Lab, ByteDance}
\icmlaffiliation{cuhk}{Department of Electronic Engineering, The Chinese University of Hong Kong}

\icmlcorrespondingauthor{Peize Sun}{peizesun@connect.hku.hk}

\icmlkeywords{Machine Learning, ICML}

\vskip 0.3in
]



\printAffiliationsAndNotice{}  

\begin{abstract}
Object detection has recently achieved a breakthrough for removing the last one non-differentiable component in the pipeline, Non-Maximum Suppression (NMS), and building up an end-to-end system. However, what makes for its one-to-one prediction has not been well understood. In this paper, we first point out that \textit{one-to-one positive sample assignment} is the key factor, while, one-to-many assignment in previous detectors causes redundant predictions in inference. Second, we surprisingly find that even training with one-to-one assignment, previous detectors still produce redundant predictions. We identify that \textit{classification cost} in matching cost is the main ingredient: (1)~previous detectors only consider location cost, (2)~by additionally introducing classification cost, previous detectors immediately produce one-to-one prediction during inference. We introduce the concept of \textit{score gap} to explore the effect of matching cost. Classification cost enlarges the score gap by choosing positive samples as those of highest score in the training iteration and reducing noisy positive samples brought by only location cost. Finally, we demonstrate the advantages of end-to-end object detection on crowded scenes. 
\vspace{-3mm}
\end{abstract}


\section{Introduction}
\label{introduction}
Object detection is one of the fundamental tasks in the computer vision area and enables numerous downstream applications. It aims at localizing a set of objects and recognizing their categories in an image. The development of object detection pipeline~\cite{RCNN,FastRCNN,FasterRCNN,CascadeRCNN,YOLO,SSD,FocalLoss,FCOS,CenterNet,DETR} is a route to remove manually-designed components and towards end-to-end system. 

\begin{figure}[t]
\includegraphics[width=0.48\textwidth]{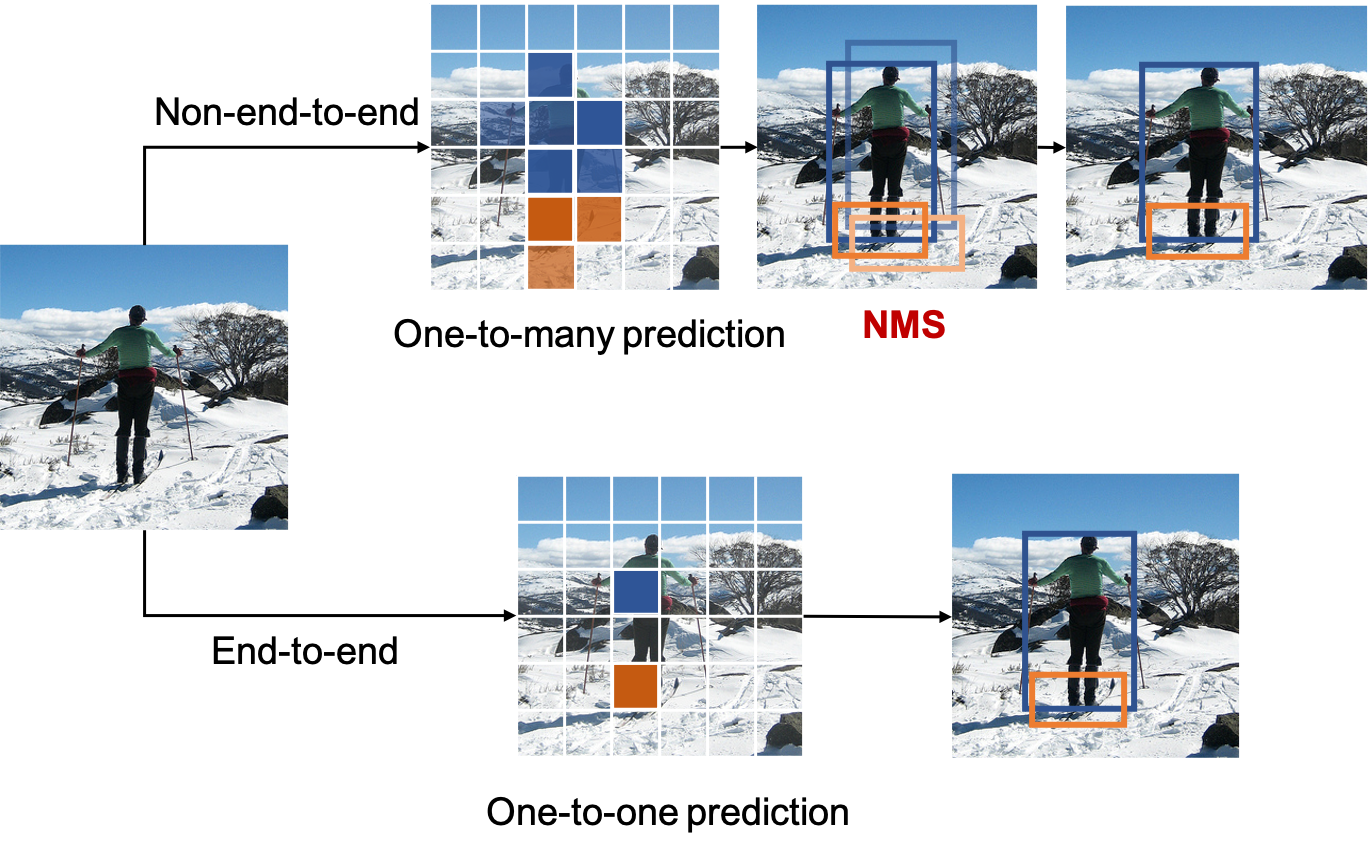}
\vspace{-5mm}
\caption{\textbf{End-to-end object detection.} Non-end-to-end object detectors require NMS to remove redundant predictions. As the last one manually-designed component in the object detection pipeline, non-differentiable NMS blocks setting up an end-to-end object detection system.}
\label{fig:end-to-end}
\vspace{-5mm}
\end{figure}

For decades, the sample in object detection is box candidates. In classical computer vision, the classifier is applied on sliding windows enumerated on the image grid~\cite{hog, dpm, boosted}. Modern detectors pre-define thousands of anchor boxes on the image grid and perform classification and regression on these candidates~\cite {RCNN, FasterRCNN, FocalLoss, YOLO9000}.

Despite box candidate methods dominate object detection for years, the detection performance is largely sensitive to sizes, aspect ratios, and the number of anchor boxes. To eliminate the hand-crafted design and complex computation of box candidates, anchor-free detectors~\cite{FCOS,CenterNet} are rising. These methods directly treat grid points in the feature map as object candidates and predict the offset from the grid point to the object box's boundaries and largely simplify the detection pipeline.

However, both box candidates and point candidates suffer from one common problem, that is, redundant and near-duplicate predictions for each object are produced, thus making non-maximum suppression(NMS) necessary post-processing in inference. Towards building up an end-to-end object detection system, NMS is the last one manually-designed component in the pipeline.

Recently, attention-based detectors~\cite{RelationNetworks,DETR,deformabledetr,sun2020sparse} achieve to directly output predictions without NMS.
Thus far, all manually-designed components in the pipeline are removed and an end-to-end object detection system is finally set up. 
However, both attention-based architecture and one-to-one positive sample assignment of these detectors are brand new compared with previous methods based on box and point candidates.
\textit{It motivates us to explore what exactly makes for end-to-end object detection.}
\footnote{The code is available at: {\url{https://github.com/PeizeSun/OneNet}}.}

In order to understand what enables non-redundant prediction in object detection, we study on three non-end-to-end detectors, RetinaNet~\cite{FocalLoss}, CenterNet~\cite{CenterNet}, FCOS~\cite{FCOS} and three end-to-end detectors, DETR~\cite{DETR}, Deformable DETR~\cite{deformabledetr}, Sparse R-CNN~\cite{sun2020sparse}. Our empirical findings show that:
\vspace{-3mm}
\begin{itemize}
    \item  Non-end-to-end detectors assigning positive samples by one-to-many groundtruth-to-samples causes redundant predictions in inference, while, end-to-end detectors are one-to-one assigning. However, even training with one-to-one assignment, non-end-to-end detectors still produce redundant predictions.  

    \item The lack of classification cost is the main obstacle to achieve one-to-one prediction: (1)~non-end-to-end detectors only consider location cost. (2)~by additionally considering classification cost, these detectors immediately produce one-to-one prediction during inference, which successfully removes NMS and achieves end-to-end detection.  
    \vspace{-3mm}
\end{itemize}

Since redundant predictions are those of high classification scores, we introduce the concept of \textit{score gap} to describe the gap between the first-highest score and the second-highest score. A sufficient requirement for end-to-end detection is that the score gap should be large enough. Assigning positive samples by only location cost cannot enlarge the score gap since it chooses positive samples as those of medium classification score in training iterations, while additionally considering classification cost leads to an enough large score gap by choosing those samples of the highest score. 
Moreover, we identify that positive samples chosen by only location cost introduce background-like positive samples, thus decrease the discriminative ability of the network, while classification cost could reduce these noisy samples.

We analyze the convergence properties of one-to-one positive sample assignment with classification cost using perceptron's update rule in the linearly separable setting.

End-to-end object detectors avoid NMS dilemma~\cite{doubleanchor} in crowded scenes. In CrowdHuman dataset~\cite{crowdhuman}, we demonstrate that end-to-end versions of RetinaNet and FCOS outperform their baseline settings by a large margin.

\section{Preliminary on Object Detection}
Object detection is a multi-task of localizing a set of objects and recognizing their categories in an image. For an input image of $H \times W \times 3$, the predictions are $N$ boxes with categories of $N \times K$ and locations of $N \times 4$, where $K$ is the number of categories and $4$ is coordinates of four sides. 

\subsection{Pipeline}

\myparagraph{Object Candidate.} Object detectors assume a region in feature map~\cite{RCNN,FasterRCNN,CascadeRCNN} or a point in feature map~\cite{YOLO,FocalLoss,FCOS,CenterNet} as the object candidate. The number of object candidates is always much more than possible objects to guarantee detection recall.

\myparagraph{Classification and Location.}
The classification sub-net predicts the probability of object candidate for $K$ object categories. The location sub-net predicts the offset from each object candidate to $4$ boundaries of the object box.

\subsection{Training} 
\myparagraph{Loss of object detection.}
The training loss of the object detection includes classification loss and regression loss, where regression loss is only executed on positive sample:

\begin{equation}
\begin{aligned}
   L \ \ &= \sum_{i \in \mathcal{P} \cup \mathcal{S}\setminus \mathcal{P}} L_{cls}(i) + \sum_{i \in \mathcal{P}}L_{loc}(i) \\
     & = \sum_{i \in \mathcal{P}}[L_{cls}(i) + L_{loc}(i)] +
   \sum_{i \in \mathcal{S}\setminus \mathcal{P}} L_{cls}(i) 
\end{aligned}
\end{equation}
where $\mathcal{S}$ is the set of samples, $\mathcal{P}$ is the set of positive sample, $\mathcal{S}\setminus \mathcal{P}$ is the set of negative sample, $L_{cls}$ is classification loss between predicted category and ground-truth category, such as cross entropy loss and Focal Loss~\cite{FocalLoss}, $L_{loc}$ is location loss between sample box and ground-truth box, such as L1 loss and GIoU loss~\cite{GIoU}. 

Though the training loss of object detection is well-defined, positive samples are controversial. In object detection, the annotation is bounding box and category of the object in the image, instead of object candidates. Selecting positive samples in object detection task is more complicated than image-level classification task, since positive samples and negative ones for image classification are indisputable when the image annotation is given. 

\myparagraph{Matching cost.} To better select positive samples and negative ones for object detection, matching cost is introduced to measure the distance between the sample and object. For sample $i$ and object $j$, the matching cost $C_{i,j}$ is:
\begin{equation}
\begin{aligned}
   C_{i,j} =  C_{cls}(i,j) + C_{loc}(i,j)
\end{aligned}
\end{equation}
where $C_{cls}(i,j)$ is classification loss between predicted category of sample $i$ and ground-truth category of object $j$, $C_{loc}(i,j)$ is location loss between sample $i$ and ground-truth box of object $j$. For convenience, we call $C_{loc}(i,j)$ as location cost, and $C_{cls}(i,j)$ as classification cost.

The matching cost is not required to be strictly equal to the loss function, as long as its design is suitable to select positive samples. In fact, the matching cost only contains $C_{loc}(i,j)$ for decades before recently~\cite{DETR,deformabledetr,sun2020sparse}.

\myparagraph{Positive sample assignment.} Once the matching cost between all samples and objects $j$ is computed, those samples below the cost threshold $\theta(j)$ will be chosen as positive samples:

\begin{equation}
\begin{aligned}
   \mathcal{P} = \{ \ i \ | \ C_{i,j} < {\rm} \theta(j), i \in \mathcal{S} \}
\end{aligned}
\end{equation}

Many heuristic rules~\cite{RCNN,CascadeRCNN,FCOS,CenterNet,ATSS,DynamicRCNN} are proposed to determine $\theta$, which lead to one-to-many and one-to-one assignment of groundtruth-to-positive samples.

\subsection{Inference.}
Since the object candidates are always much more than objects in the image, the output is filtered by the score threshold to guarantee detection precision. If there remain redundant boxes, non-maximum suppression(NMS) is used to remove these redundant predictions. NMS is a heuristic manually-designed component. The box with the maximum score is selected and others neighboring boxes are eliminated. 

However, non-differentiable NMS blocks the establishment of an end-to-end system. Worsely, detectors suffer from NMS dilemma in crowded scene~\cite{doubleanchor}. To this end, end-to-end object detection is proposed.

\begin{figure}[t]
\centering
\includegraphics[width=0.45\textwidth]{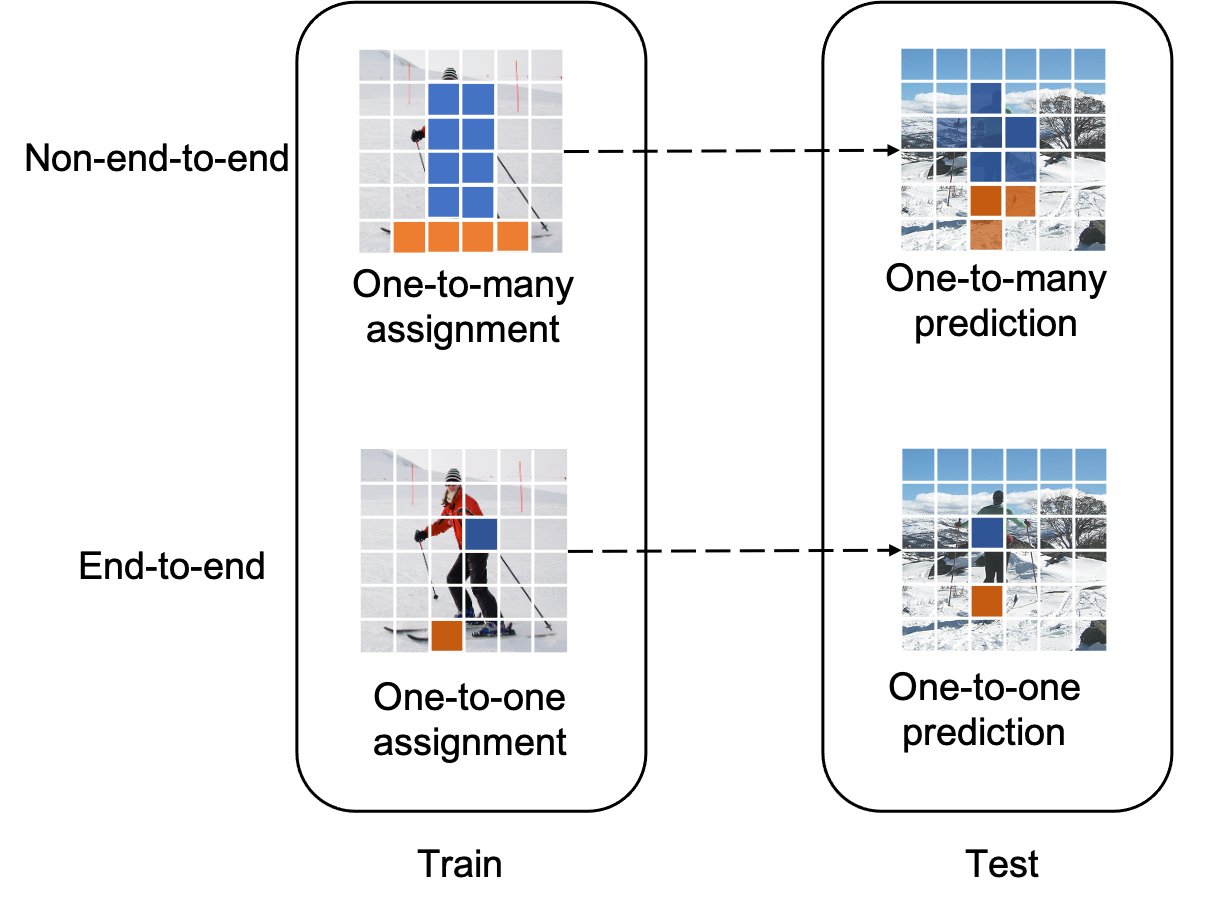}
\vspace{-5mm}
\caption{\textbf{Positive sample assignment.} Non-end-to-end detectors apply one-to-many positive sample assignment in training and produce one-to-many predictions in inference. While, end-to-end object detectors are one-to-one positive sample assignment and one-to-one prediction. This motivates us to apply one-to-one assignment in non-end-to-end detectors.}
\label{fig:one2one}
\vspace{3mm}
\end{figure}

\begin{table}[t!]
\vspace{-3mm}
\begin{center}
\setlength{\tabcolsep}{3.2mm}
\begin{tabular}
{l c c l}
\toprule
Detector & o2o& 
\multicolumn{1}{c}{AP} & 
\multicolumn{1}{c}{AP(+NMS)} \\
\midrule
DETR& 
\cellcolor{lightgray}\checkmark&
\cellcolor{lightgray}40.0 & 
\cellcolor{lightgray}39.9 (\textcolor{blue}{-0.1})\\ 
Deformable DETR& 
\cellcolor{lightgray}\checkmark&
\cellcolor{lightgray}44.0 & 
\cellcolor{lightgray}43.9 (\textcolor{blue}{-0.1})\\ 
Sparse R-CNN& 
\cellcolor{lightgray}\checkmark&
\cellcolor{lightgray}45.0 & 
\cellcolor{lightgray}44.9 (\textcolor{blue}{-0.1})\\ 
\midrule
RetinaNet&
\cellcolor{lightgray}&
\cellcolor{lightgray}7.7& 
\cellcolor{lightgray}37.4 (\textcolor{blue}{+29.7})\\
 & \checkmark & 33.6 & 36.8 (\textcolor{blue}{+3.2})\\
CenterNet& 
\cellcolor{lightgray}&
\cellcolor{lightgray}24.9 & 
\cellcolor{lightgray}35.0 (\textcolor{blue}{+10.1})\\
 & \checkmark & 23.4 & 32.0 (\textcolor{blue}{+8.6}) \\
FCOS&
\cellcolor{lightgray}&
\cellcolor{lightgray}17.3& 
\cellcolor{lightgray}38.7 (\textcolor{blue}{+21.4})\\
 & \checkmark & 34.9 & 37.7 (\textcolor{blue}{+2.8})\\
\bottomrule
\end{tabular}
\end{center}
\vspace{-3mm}
\caption{\textbf{Effect of one-to-one positive sample assignment.} The detectors' original settings are highlighted by gray. ``o2o'' means one-to-one positive sample assignment. The top section is end-to-end detectors, which apply one-to-one assignment and don't depend on NMS. The bottom section is non-end-to-end detectors, whose original settings use one-to-many assignment and heavily rely on NMS. Training with one-to-one assignment only reduces non-end-to-end detectors' dependence on NMS to some extent, they still need NMS to further remove redundant predictions.}
\label{table:one2one}
\end{table}

\section{End-to-End Object Detection} \label{sec:end-to-end}
End-to-end object detection means that object detection pipeline is \textit{without any non-differentiable component, \eg, NMS}. The input of network is the image and the output is direct predictions of classification on object categories or background and the box regression. The whole network is trained in an end-to-end manner with back-propagation.

\subsection{Experiment Setting}

\myparagraph{Detectors.}
We select three non-end-to-end detectors, RetinaNet~\cite{FocalLoss}, CenterNet~\cite{CenterNet}, FCOS~\cite{FCOS} and three end-to-end detectors, DETR~\cite{DETR}, Deformable DETR~\cite{deformabledetr}, Sparse R-CNN~\cite{sun2020sparse}.

\myparagraph{Dataset.}
Our experiments are conducted on the challenging COCO benchmark~\cite{COCO}. We use the standard COCO metrics AP of averaging over IoU thresholds. All models are trained on \texttt{train2017} split ($\sim$118k images) and evaluated with \texttt{val2017} (5k images).

\subsection{Positive Sample Assignment}

\myparagraph{One-to-many assignment.} 
The remarkable property of non-end-to-end detectors is one-to-many positive sample assignment, as shown in Figure~\ref{fig:one2one}. In the training step, for one ground-truth box, any sample whose matching cost is below the cost threshold is assigned as the positive sample. It always causes multiple samples in the feature maps to be selected as positive samples. As a result, in the inference step, these detectors produce redundant predictions.

\myparagraph{One-to-one assignment.} On the contrary, end-to-end detectors apply one-to-one assignment during the training step. 
For one ground-truth box, only one sample with the minimum matching cost is assigned as the positive sample, others are all negative samples. The positive sample is usually selected by bipartite matching~\cite{kuhn1955hungarian} to avoid sample conflict, \ie, two ground-truth boxes share the same positive sample.

As shown in Table~\ref{table:one2one}, end-to-end detectors, including DETR, Deformable DETR and Sparse R-CNN, apply one-to-one assignment and eliminate NMS. Therefore, an intuitive idea to transform non-end-to-end detectors become end-to-end is to replace one-to-many assignment with one-to-one assignment. Specifically, RetinaNet chooses the positive sample as the anchor that has largest IoU with the ground-truth box, CenterNet chooses the grid point in the feature map that has the nearest distance to the ground-truth box center, while FCOS chooses from the pre-defined layer in feature pyramids~\cite{FCOS}.

However, one-to-one assignment only reduces the dependence on NMS to some extent, non-end-to-end detectors still need NMS to further remove redundant predictions. For example, NMS could further improve one-to-one assignment version of RetinaNet, CenterNet and FCOS by 3.2 AP, 8.6 AP and 2.8 AP, respectively, as shown in Table~\ref{table:one2one}.

\vspace{3mm}
\textbf{Conclusion 3.1} \textit{Even replacing one-to-many assignment to one-to-one assignment in training, non-end-to-end detectors still produce redundant predictions in inference.}
\vspace{3mm}

Experiments on positive sample assignment demonstrate that one-to-one assignment is necessary but not sufficient for end-to-end object detection. We further delve into the compositions of matching cost.

\begin{figure}[t]
\centering
\includegraphics[width=0.48\textwidth]{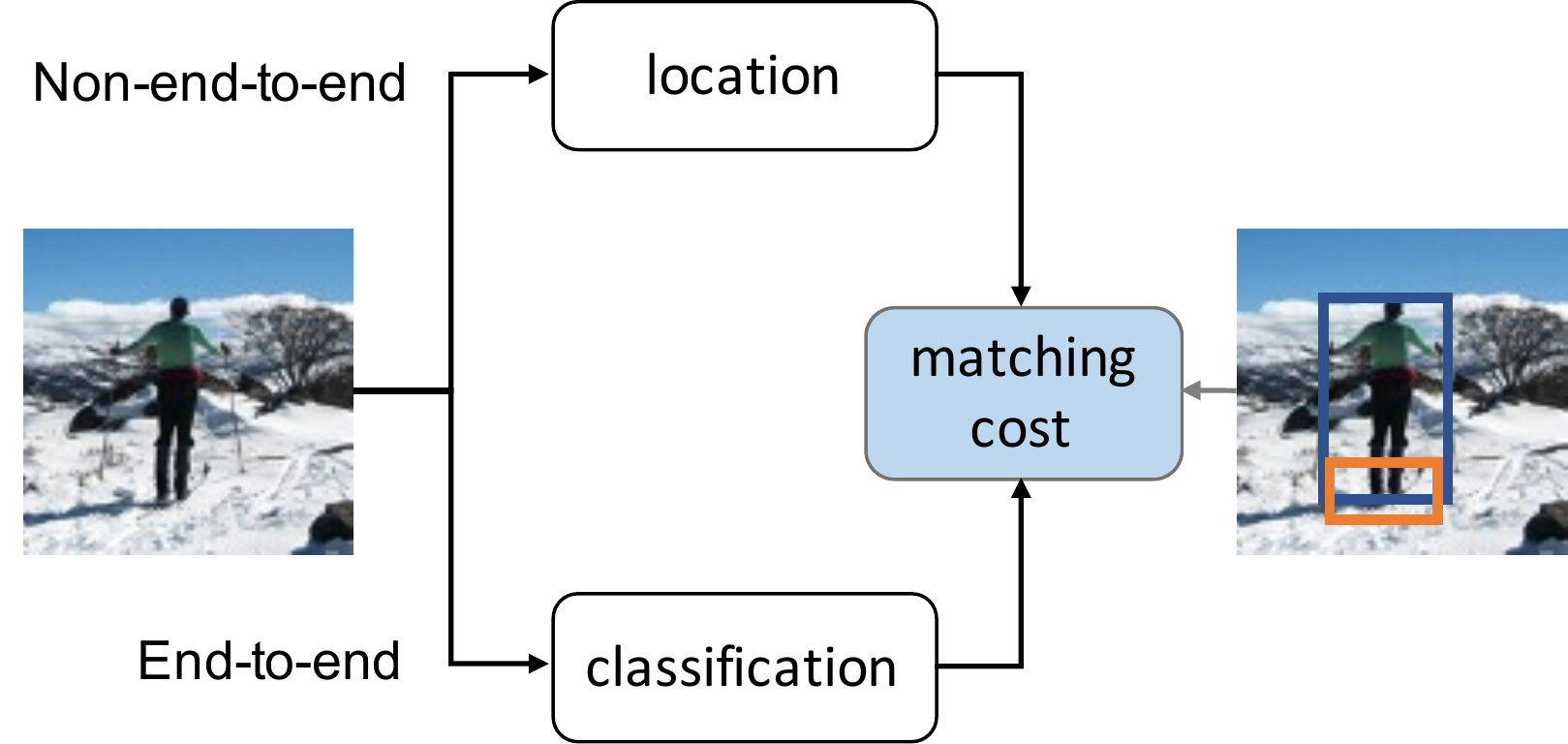}
\vspace{-3mm}
\caption{\textbf{Matching cost.} Non-end-to-end object detectors assign positive samples by only location cost, while end-to-end detectors additionally consider classification cost.}
\label{fig:matchingcost}
\vspace{4mm}
\end{figure}

\begin{table}[t!]
\vspace{-3mm}
\begin{center}
{
\setlength{\tabcolsep}{0.7mm}
\begin{tabular}
{l c c c c l}
\toprule[1.2pt]
Detector &
\multicolumn{2}{c}{loc.} &
cls. & 
\multicolumn{1}{c}{AP} & 
AP(+NMS)\\
& pre-def.\textcolor{white}{a} & pred. & & & \\
\midrule

DETR&&\checkmark  &  &
12.6 &  
23.6 (\textcolor{blue}{+11.0})\\
 &&  
\cellcolor{lightgray}\checkmark  & 
\cellcolor{lightgray} \checkmark & 
\cellcolor{lightgray}40.0 & 
\cellcolor{lightgray}39.9 (\textcolor{blue}{-0.1})\\

Deformable DETR&&\checkmark  &  &
12.0 &  
23.8 (\textcolor{blue}{+11.8})\\
 &&  
\cellcolor{lightgray}\checkmark  & 
\cellcolor{lightgray} \checkmark & 
\cellcolor{lightgray}44.0 & 
\cellcolor{lightgray}43.9 (\textcolor{blue}{-0.1})\\

Sparse R-CNN && \checkmark &  & 
20.1 & 
33.1 (\textcolor{blue}{+13.0})\\
 &&
\cellcolor{lightgray}\checkmark  & 
\cellcolor{lightgray} \checkmark & 
\cellcolor{lightgray}45.0 & 
\cellcolor{lightgray}44.9 (\textcolor{blue}{-0.1})\\ 

\midrule
RetinaNet + o2o & \checkmark & & & 33.6 & 36.8 (\textcolor{blue}{+3.2})\\
 & \checkmark &  & \checkmark& 36.0 & 36.2 (\textcolor{blue}{+0.2}) \\
 && \checkmark  & \checkmark & 37.5 & 37.5 (\textcolor{blue}{+0.0})\\
\midrule

CenterNet + o2o & \checkmark & & &23.4 & 32.0 (\textcolor{blue}{+8.6})\\
 & \checkmark &  & \checkmark& 33.3 & 33.2 (\textcolor{blue}{-0.1})\\
&& \checkmark    &  \checkmark & 34.9 & 34.8 (\textcolor{blue}{-0.1})\\
\midrule

FCOS + o2o & \checkmark & & &34.9 & 37.7 (\textcolor{blue}{+2.8})\\
 & \checkmark &  & \checkmark& 35.9 & 36.1 (\textcolor{blue}{+0.2}) \\
&& \checkmark  & \checkmark & 38.9 & 38.9 (\textcolor{blue}{+0.0})\\

\bottomrule[1.2pt]
\end{tabular}}
\end{center}
\vspace{-3mm}
\caption{\textbf{Effect of classification cost.} The detectors' original settings are highlighted by gray. ``o2o'' means one-to-one positive sample assignment. ``loc.'' means location cost. ``cls.'' means classification cost. ``pre-def.'' and ``pred.'' are pre-defined location cost and predicted location cost, illustrated in~\ref{matchingcost}. All detectors apply one-to-one positive sample assignment. Without classification cost, all detectors significantly drop the detection accuracy and heavily rely on NMS. Instead, adding classification cost eliminates the necessity of NMS.}
\label{table:matchingcost}
\vspace{3mm}
\end{table}

\subsection{Matching Cost}
\label{matchingcost}
\myparagraph{Location cost.} 
By reviewing non-end-to-end object detectors,  we identity that they assign positive samples by only  location cost. The location cost is defined as follows:
\begin{equation}
    \label{loc_cost}
    C_{loc} 
    = 
    \lambda_{iou} \cdot C_{\mathit{iou}}
    +
    \lambda_{L1} \cdot C_{\mathit{L1}} 
\end{equation}
where  $C_{\mathit{L1}}$ and $C_{\mathit{iou}}$ are L1 loss and IoU loss between sample and ground-truth box, respectively. 
$\lambda_{L1}$ and $\lambda_{iou}$ are coefficients. 
When object candidates are points in the feature map, $\lambda_{iou}=0$.  We note that object candidates could be pre-defined or predicted. Take an example of RetinaNet, its pre-defined object candidates are anchor boxes, while its predicted object candidates are predicted boxes refined by the predicted offsets. As for CenterNet and FCOS, the pre-defined object candidates are grid points in the feature map, while the predicted object candidates are predicted boxes. Based on object candidates, the location cost could also be pre-defined or predicted.

Location cost can reasonably measure whether the selected positive sample is beneficial for location. However, object detection is a multi-task of location and classification. Classification cost is supposed to be considered as well, although it has been ignored for decades before recently.

\myparagraph{Classification cost.} 
By introducing classification cost into assignment, the total cost is the summation of classification cost and location cost between sample and ground-truth, defined as follows:
\begin{equation}
    \label{total_loss}
    C = \lambda_{cls}  \cdot  C_{\mathit{cls}} + C_{\mathit{loc}}
\end{equation}
where  $C_{\mathit{cls}}$ is classification loss of predicted classifications and ground truth category labels.  $C_{\mathit{loc}}$ is defined in Equation~\ref{loc_cost}. $\lambda_{cls}$ is coefficient.

As shown in Table~\ref{table:matchingcost}, the default settings of end-to-end object detectors include both location cost and classification cost. When discarding classification cost, these detectors significantly degenerate and heavily rely on NMS. 

Continuing on one-to-one assignment versions of RetinaNet, CenterNet and FCOS, classification cost is additionally introduced to their matching cost. For RetinaNet and CenterNet, the positive sample is selected as the sample with minimum matching cost among all samples. For FCOS, the positive sample is chosen from the pre-defined layer in feature pyramids. As shown in Table~\ref{table:matchingcost}, adding classification cost immediately makes NMS has little effect on the detection performance. 

\vspace{3mm}
\textbf{Conclusion 3.2} \textit{Non-end-to-end detectors assign positive samples by only location cost. However, when additionally considering classification cost, they immediately produce one-to-one prediction under one-to-one assignment.}
\vspace{3mm}

To completely reduce the necessity of NMS, we also carry out the experiments in which the pre-defined location cost in RetinaNet, CenterNet and FCOS is changed to predicted location cost. We note that the location cost in DETR, Deformable DETR and Sparse R-CNN is also based on predicted boxes. As shown in Table~\ref{table:matchingcost}, the combination of classification cost and predicted location cost enable previous non-end-to-end detectors to achieve completely end-to-end. Interestingly, the location cost based on predicted boxes could obtain better detection performance. We explain it is because predicted location cost makes matching cost more aligned to the training loss function, thus benefits to the optimization of the object detector.

Our experiments above demonstrate that one-to-one assignment is necessary but not sufficient for one-to-one prediction. Additionally considering classification cost is the key to achieve end-to-end object detection. We further explore how classification cost makes an effect.

\subsection{Score Gap}
In order to understand how classification cost contributes to end-to-end object detection, we first introduce the following definition.

\begin{figure}[t]
\centering
\includegraphics[width=0.45\textwidth]{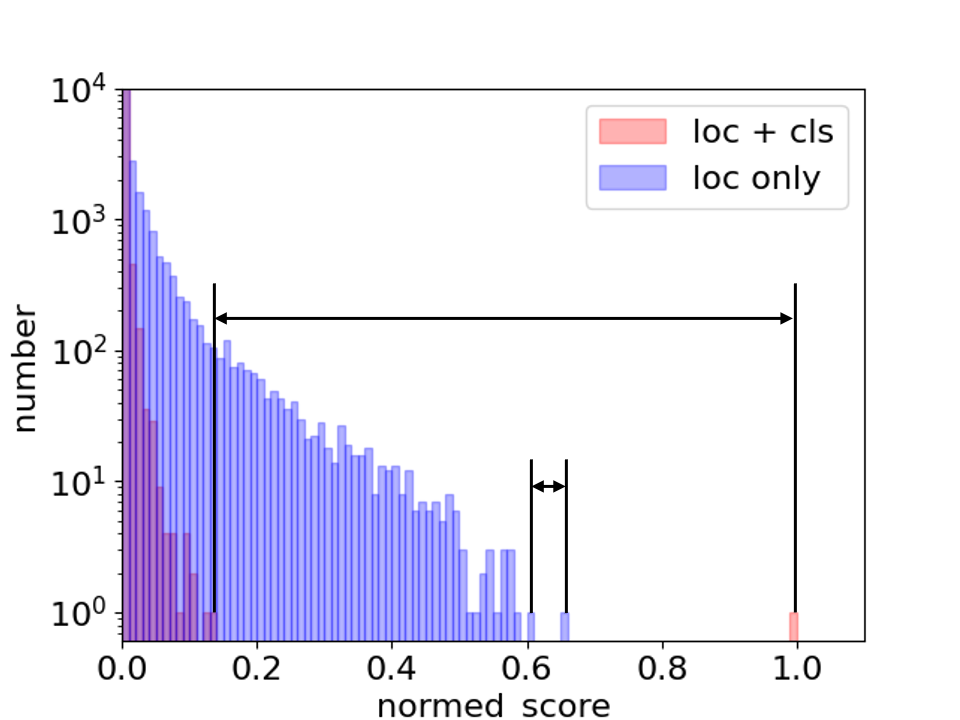}
\caption{\textbf{Samples' classification scores of the trained detector.} For better visualization, we only show the part below number of $10^4$, and scores are normalized to [0, 1]. Blue bins show the detector trained with positive samples chosen by only location cost. Red bins consider both location cost and classification cost. Classification cost results in a clear score gap between samples of first highest score and second highest score.}
\label{fig:margin}
\vspace{5mm}
\end{figure}

\begin{figure}[!t]
\hspace{-2mm}
\begin{subfigure}{0.51\textwidth}
     \includegraphics[width=1.00\textwidth]{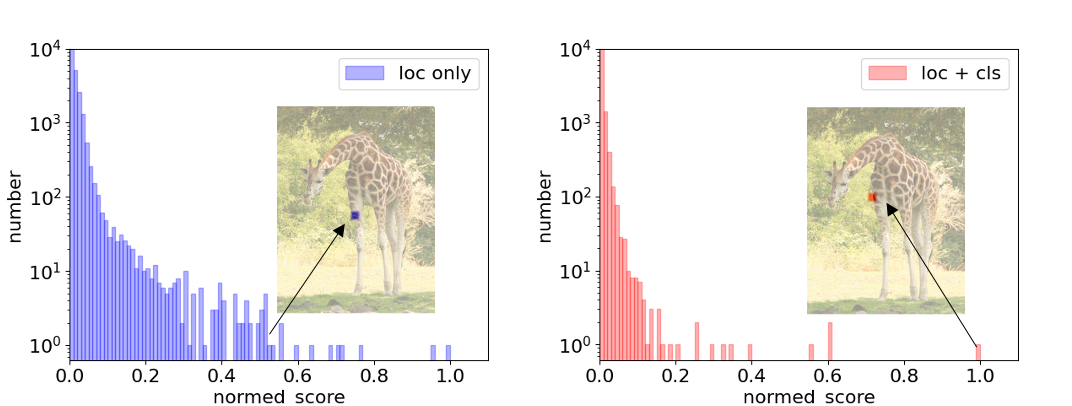}
     \caption{Training early stage.}
     \label{fig:epocha}
    \vspace{1mm}
\end{subfigure}

\hspace{-2mm}
\begin{subfigure}{0.51\textwidth}
     \includegraphics[width=1.00\textwidth]{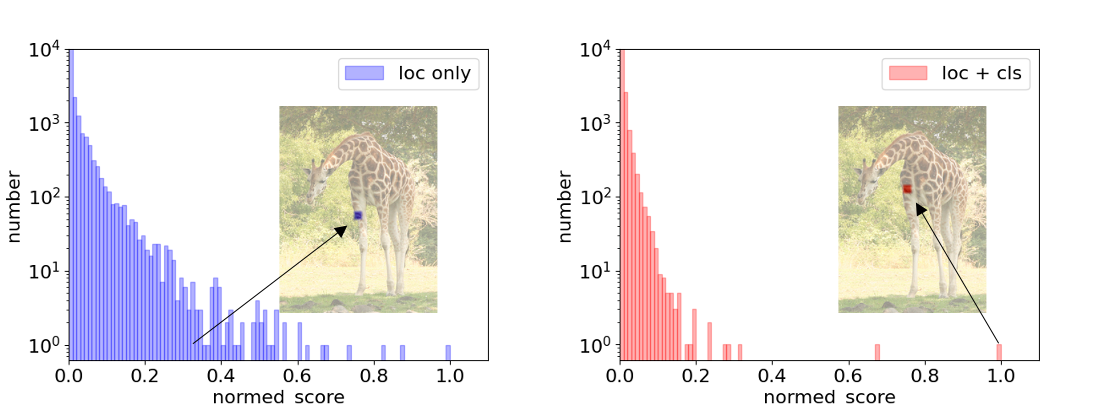}
     \caption{Training middle stage.}
     \label{fig:epochb}
    \vspace{1mm}
\end{subfigure}

\hspace{-2mm}
\begin{subfigure}{0.51\textwidth}
     \includegraphics[width=1.00\textwidth]{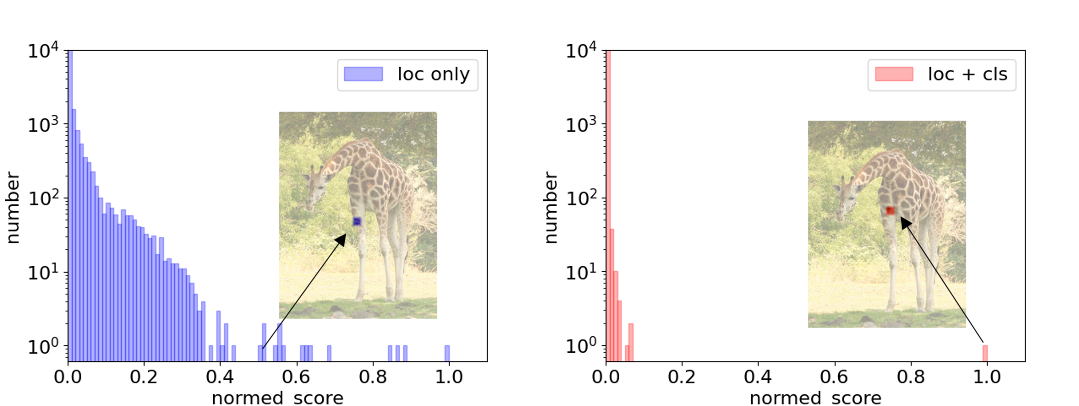}
     \caption{Training late stage.}
     \label{fig:epochc}
    \vspace{1mm}
\end{subfigure}

\vspace{2mm}
\caption{\textbf{Positive samples in different training stages.} For better visualization, we only show the part below the number of $10^4$, and scores are normalized to [0, 1]. Blue bins show the detector trained with positive samples chosen by only location cost. Red bins consider both location cost and classification cost. Only location cost selects the positive samples as those of medium score. Introducing classification loss makes positive samples as those of the highest score during the whole training process. 
}
\label{fig:epoch}
\vspace{3mm}
\end{figure}

\textbf{Definition 3.3 (Score Gap)} \textit{Given a classification network $\mathcal{N}$ and a set of samples $\mathcal{S}$, if sample $i$ is positive and others are negative, train the network and get each sample's score s(i), let $i_{max} = {\rm arg}\max_{j \in \mathcal{S}}s(j)$, then score gap ($\mathcal{N}$,  $\mathcal{S}$, $i$) is defined as:}

\begin{equation}
\begin{aligned}	
	{\rm score \ gap}( \mathcal{N},  \mathcal{S}, i) = \min_{j \in \mathcal{S} \setminus i_{max} } ( s(i_{max}) – s(j) )
\end{aligned}
\end{equation}

The score gap describes the gap between the first-highest score and the second-highest score. A sufficient requirement for end-to-end object detection is that the score gap should be large enough, otherwise, non-maximum predictions cannot be easily filtered out: a high score threshold may filter out all predictions, while a low threshold may output redundant predictions. 

In Figure~\ref{fig:margin}, we show samples' classification scores from the trained detectors with and without classification cost under one-to-one assignment. For only location cost, the gap between the highest score and the second-highest score is negligible. Also, all samples are relatively lower scores. Instead, considering classification cost produces a clear score gap, therefore, achieves end-to-end object detection.

To explore how score gap is produced under different matching costs, we further show samples' classification scores during different training stages in Figure~\ref{fig:epoch}. 

For only location cost, the positive sample lies in the grid point closest to the center of the object ground-truth box. Nevertheless, the positive samples are those of medium score. Such positive samples will push the network to pull down the score of samples that have been high score. As a consequence, all samples tend to be relatively lower scores.

When additionally considering classification cost, the positive sample is those of highest score in the training iterations. These choices are much more useful to further increase the score of positive samples and widen the score gap, meanwhile, it does not hurt the box regression since the positive samples are still inside the object ground-truth box. After the whole training process, a large enough score gap is finally generated to achieve end-to-end object detection.

\vspace{3mm}
\textbf{Conclusion 3.4} \textit{Classification cost chooses positive samples as those of highest score in the training process, therefore, produces large enough score gap for end-to-end object detection.}
\vspace{3mm}

\begin{figure*}[t]
\centering
\includegraphics[width=1.00\textwidth]{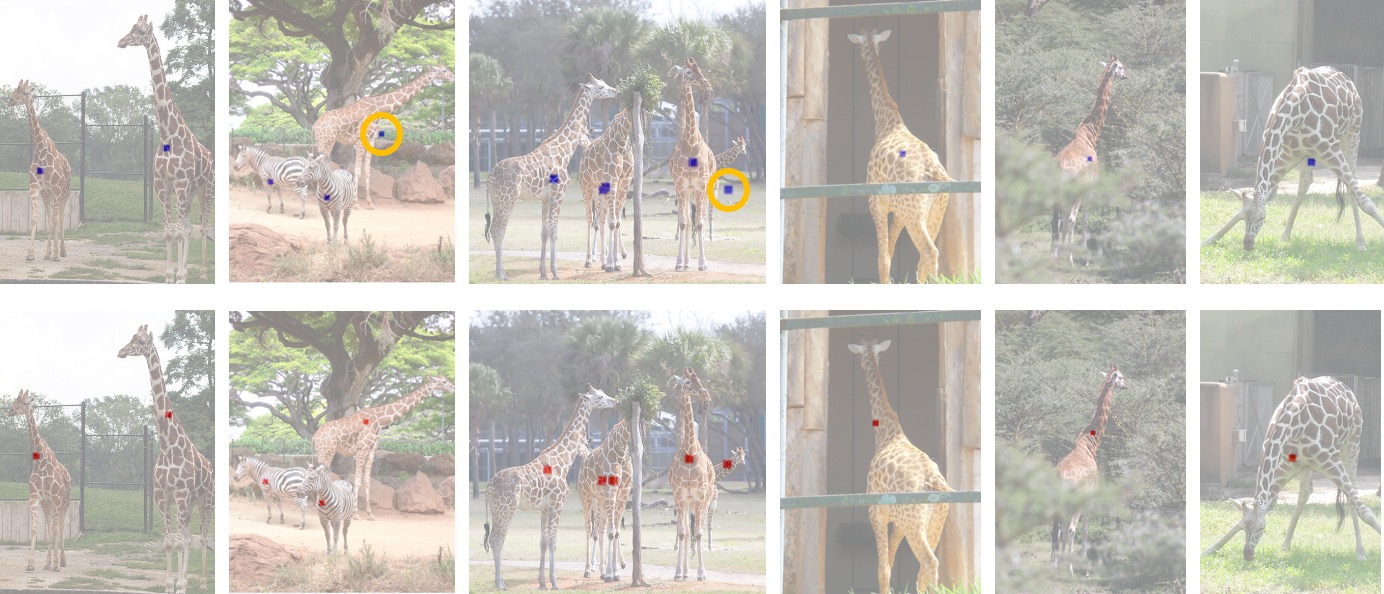}
\caption{\textbf{Positive samples in different training images.} For better visualization, the positive grid points are highlighted by surrounding circles. 1st row is only location cost. 2nd row is the summation of classification cost and location cost.  The positive samples assigned by only location cost are the grid point closest to the ground-truth box center, however, some background-like samples are assigned as positive samples, highlighted by yellow rings. Adding classification cost, positive samples are grid points in more discriminative areas, \eg, neck of the giraffe.}
\label{fig:pos}
\vspace{2mm}
\end{figure*}

We note that the positive sample selected by only location cost is the same sample during the whole training process, but classification score of this sample is always kept as the medium score. To explain why its score can't be lifted, we visualize positive samples in different training images, as shown in Figure~\ref{fig:pos}.

If only considering location cost, the positive sample lies in the grid point closest to the center of the object ground-truth box. This assignment is beneficial for box regression, but is not a good choice for foreground and background classification. Specifically, some background-like samples are assigned as positive samples, highlighted by yellow rings in Figure~\ref{fig:pos}. These cases come from objects' arbitrary shapes and poses, such as the long neck of the giraffe. These background-like samples are noisy samples for classification task and decrease the discriminative ability of the network.

On the contrary, when classification cost is introduced, positive samples are grid points in more discriminative areas, \eg, neck of the giraffe. In such cases, it is avoided to select positive samples outside the area of the object. Moreover, these discriminative positive samples are also more useful for classification branch to distinguish noisy samples. As a result, noisy samples are effectively reduced from positive samples.

\vspace{3mm}
\textbf{Observation 3.5 } \textit{Only location cost may select noisy background-like positive samples, while additionally considering classification cost could reduce these noisy positive samples.}
\vspace{3mm}

From the above analysis, we discover that non-end-to-end detectors only consider location cost to select positive samples, which makes noisy positive samples and decreases the discriminative ability of the network. This leads to a small score gap and produces redundancy predictions. Instead, when classification cost is additionally introduced, the noisy samples could be reduced, the score gap is large enough, therefore, end-to-end object detection is achieved.

\section{Theoretical Analysis}
\label{sec:theoretical}
\subsection{Setup}

In this section, we analyze the convergence properties of object detectors under one-to-one assignment with matching cost as the summation of location cost and classification cost, in which only one sample with the minimum matching cost is assigned as the positive sample, others are all negative samples.

Since a systematic framework is beyond our reach, we first make some reasonable assumptions based on verification experiments. We conduct the experiment in which the parameter of location sub-net is fixed, only classification sub-net is trained. And we observe the same conclusions as Section~\ref{sec:end-to-end}. This leads to the following observation:

\vspace{2mm}
\textbf{Observation 4.1 } \textit{The optimization of classification sub-net is irrelevant to location sub-net. } 

Based on Observation 4.1, analysis of classification score of object detection can be reasonably simplified as a single classification problem, in which only the sample with minimum classification cost is chosen as the positive sample among all samples, others are all negative samples.

We focus on analyzing properties using linear classifier. Let $\mathcal{X}=\{x\in\mathbb{R}^d : \left\|x\right\| \leq 1 \}$ 
be an instance space and $\mathcal{Y} = \{+1, -1\}$ be the label space. 
The label of a positive sample is $+1$ while the label of a negative sample is 
$-1$. We wish to train a classifier $h$, coming from a hypothesis class 
$\mathcal{H}=\{x \mapsto \mathrm{sign}( w^T x) : w\in\mathbb{R}^d\}$. 
Note that we can express the bias term $b$ by rewriting 
$w=[\hat{w},b]$ and $x=[\hat{x}, 1]$. We use the perceptron’s 
update rule with mini-batch size of 1. That is, given the classifier 
$w_t\in\mathbb{R}^d$, the update is only performed on incorrectly 
classified example $(x_t,y_t)\in \mathcal{X} \times \mathcal{Y}$ as given 
by $w_{t+1} = w_t +\eta y_tx_t$ where $\eta$ is the stepsize. According to one-to-one positive label assignment, in each update step, we denote $x_t^1 =\arg\max_{x\in\mathcal{X}} w_t\tran x$, the label of $x_t^1$ is $y(x_t^1)=+1$ and the labels of the remaining samples in $\mathcal{X}$ are $y(x)=-1, x\in\mathcal{X}\backslash \{x_t^1\}$.

\subsection{Theoretical Results}

We first show that samples with labels assigned by one-to-one assignment are linearly separable at each training iteration, implying the positive definite score gap. Based on this result, then we show that the one-to-one assignment can converge within finite update steps. 

\textbf{Proposition 4.2 (Feasibility)} \textit{
Suppose that the one-to-one assignment is run on a sequence of examples 
from $\mathcal{X} \times \mathcal{Y}$. Given weight vector $w_t = 
[\hat{w}_{t}, b_{t}]$ at update step $t$, there exists $\gamma_t\in 
\mathbb{R}$ and $\delta_t >0$ such that for all $(x,y) \in \mathcal{X} 
\times \mathcal{Y}$ we have $y(w^*_t)^T x \geq \delta_t$ with $w_t^* =
[\hat{w}_t, \gamma_t]$.}

\textit{Proof.} Detailed proof is provided in Appendix.

By Proposition 4.2, we see that there always exists a classifier that can correctly classify all samples at every update step when the label is assigned by one-to-one assignment.

\vspace{2mm}
\textbf{Theorem 4.3 (Convergence)}
\textit{ Let $\gamma_{t+1}$ and $\gamma_t$ be the constants defined in Proposition 4.2. For each update step $t$, we assume there exists a stepsize $\eta_t$ such that $\left\|x_t\right\|^2\eta_t^2+y_t(\gamma_{t+1}-2\gamma_{t})\eta_t+b_t(\gamma_{t+1}-\gamma_{t})>0$ where $(x_t, y_t)$ be the incorrectly classified sample at iteration $t$. 
If the sample label is assigned by one-to-one assignment, then, $t \leq \frac{{\eta_{max}^2}-2\eta_{min}\delta_{\min}(w_{1}\tran w_{0}^*-\left\|w_{0}\right\| -{\eta_{max}})}{2\eta_{min}^2\delta_{\min}^2}$ where $\eta_{max}$ and $\eta_{min}$ are the maximum and the minimum value of stepsize among all $t$'s updates, $w_1$  is the classifier after the first update and $\delta_{\min}$ is the minimum of all $\delta_t$s in Proposition 4.2. All instances at initialization can be correctly classified by $w_0^*$.
} 

\textit{Proof.} Detailed proof is provided in Appendix.

Theorem 4.3 shows that samples with labels assigned by one-to-one assignment can converge to a classifier that admits a single positive sample, \ie, its label is $+1$. Therefore it is guaranteed to converge to a solution in the sense that the classification output is one-to-one prediction.

\vspace{2mm}
\textbf{Remark 4.4} \textit{
The one-to-one assignment with classification cost is guaranteed to converge to a solution in the sense that the classification output is one-to-one prediction. But assignment with only location cost may produce multiple positive samples.}

By Theorem 4.3, we see the one-to-one prediction is based on that there exists a classifier that can correctly classify all samples at every update step. However, without classification cost, only location cost determines positive samples by location criterion, which may cause the problem that such the positive sample may not be linearly separable with the remaining negative samples. In this case, perception learning algorithm can converge to a classifier that makes the fewest errors in prediction \cite{burton1997perceptron}. Hence, it is likely to produce many positive samples.

\section{Crowded Object Detection}
\label{sec:crowd}
In crowded scenarios, previous non-end-to-end detectors suffer from one dilemma when using NMS to remove duplicate predictions~\cite{doubleanchor}: higher NMS threshold brings more false positives, while a lower threshold may mistakenly remove true positives and cause undetected objects. On the contrary, end-to-end detectors completely avoid this problem by eliminating NMS, and exhibit superior performance in crowded scenes.

\subsection{Experiment Setting}

\myparagraph{Detectors.} 
We select RetinaNet~\cite{FocalLoss}, FCOS~\cite{FCOS} and their end-to-end variants with predicted location cost.

\myparagraph{Dataset.}
CrowdHuman~\cite{crowdhuman} is a widely-used benchmark for crowded object detection, in which human boxes are highly crowded and overlapped. We use metrics AP, mMR and recall of on IoU 0.5 threshold. All models are trained on training set ($\sim$15k images) and evaluated with validation set ($\sim$4k images).

\begin{table}[t]
\begin{center}
\setlength{\tabcolsep}{0.5mm}
\begin{tabular}
{
l | c | c c c 
}
\toprule
Method & 
NMS &
AP$_{50}$   & 
mMR$\downarrow$  & 
Recall \\
\midrule

Annotation~\cite{defcn} & \checkmark &  - & - & 95.0 
\\
\midrule

RetinaNet~\cite{FocalLoss} & \checkmark & 81.7 & 57.6 & 88.6 
\\
RetinaNet + o2o + cls. & $\circ$ & 90.8 & 49.3 & 98.1 
\\
& \checkmark & 86.3 & 49.9 & 93.2 
\\
\midrule

FCOS~\cite{FCOS} & \checkmark &  86.1 & 55.2 & 94.3 
\\

FCOS + o2o + cls. & $\circ$ & 90.7 & 48.2 & 97.6 
\\
& \checkmark &  86.0 & 49.0 & 92.3 
\\
\bottomrule
\end{tabular}
\end{center}
\vspace{-4mm}
\caption{\textbf{Comparisons with different object detectors on CrowdHuman validation set.} ``$\circ$” means no NMS processing. Annotation boxes processed by NMS only obtain 95.0\% recall, which is the upper bound of non-end-to-end detectors. End-to-end versions of RetinaNet and FCOS are not constrained to that recall upper bound and outperform their baselines setting by a large margin. NMS damages the performance of end-to-end detectors on crowded scenes.}
\label{table:crowd}
\vspace{-2mm}
\end{table}

\subsection{Results}
Table~\ref{table:crowd} shows the performance of different object detectors on CrowdHuman. We first show that applying NMS on annotation boxes only obtains 95\% recall, which indicates the even strongest non-end-to-end detectors are bounded to NMS in crowded scenes. Instead, when we reform RetinaNet~\cite{FocalLoss} and FCOS~\cite{FCOS} to end-to-end detectors by adding one-to-one positive sample and classification cost, they are not constrained to this recall upper bound and significantly improve the recall to 98.1\% and 97.6\%, respectively. Meanwhile, AP$_{50}$ and mMR benefit a large improvement from end-to-end setting. When NMS is used to process the predictions of end-to-end RetinaNet and FCOS, the performance degenerates at once. It furthermore demonstrates the disadvantage of NMS in crowded scenes and the superiority of end-to-end object detection.

\section{Related Work}
\myparagraph{Object detection.} Object detection is one of the most fundamental and challenging topics in computer vision fields. Limited by classical feature extraction techniques~\cite{hog, boosted}, the performance has plateaued for decades, and the application scenarios are limited. With the rapid development of deep learning~\cite{AlexNet,vgg,GoogLeNet,ResNet,DenseNet}, object detection achieves powerful performance~\cite{PASCAL-VOC,COCO}.

\myparagraph{One-stage detector.}
One-stage detector directly predicts the category and location of dense anchor boxes or points over different spatial positions and scales in a single-shot manner such as YOLO~\cite{YOLO},
SSD~\cite{SSD} and RetinaNet~\cite{FocalLoss}. 
YOLO~\cite{YOLO} divides the image into an S × S grid, and if the center of an object falls into a grid cell, the corresponding cell is responsible for detecting this object. SSD~\cite{SSD} directly predicts object category and anchor box offsets on multi-scale feature map layers. RetinaNet~\cite{FocalLoss} utilizes focal loss to ease the extreme unbalance of positive and negative samples based on the FPN~\cite{FPN}. 
Recently, anchor-free detectors~\cite{DenseBox} is proposed to make this pipeline much simpler by replacing hand-crafted anchor boxes with reference points. CornerNet~\cite{CornerNet} generates the keypoints by heatmap and group them by the Associative Embedding~\cite{associative}. CenterNet~\cite{CenterNet} directly uses the center point to regress the target object on a single scale. FCOS~\cite{FCOS} assigns the objects of different size and scales to multi-scale feature maps with the power of FPN~\cite{FPN}. ATSS\cite{ATSS} reveals that the essential difference between anchor-based and anchor-free detection is how to define positive and negative training samples, leading to the performance gap between them. 

\myparagraph{Two-stage detector.}
The two-stage detectors~\cite{CascadeRCNN, R-FCN, FastRCNN, MaskRCNN, FasterRCNN} firstly generate a high-quality set of foreground proposals by region proposal networks and then refine each proposal's location and predicts its category. Fast R-CNN~\cite{FastRCNN} uses Selective Search~\cite{SelectiveSearch} to generate foreground proposals and refine the proposals in R-CNN~\cite{RCNN} Head. Faster R-CNN ~\cite{FasterRCNN} proposes the region proposal network, which generates high-quality proposals in real-time. Cascade R-CNN~\cite{CascadeRCNN} iteratively uses multiple R-CNN heads with different label assign threshold to get high-quality detection boxes. Cascade RPN~\cite{CascadeRPN} improves the region proposal quality and detection performance by systematically addressing the limitation of the conventional RPN that heuristically defines the anchors and aligns the features to the anchors. Libra R-CNN~\cite{LibraRCNN} tries to solve the unbalance problems in sample level, feature level, and objective level. Grid R-CNN~\cite{GridRCNN} adopts a grid-guided localization mechanism for accurate object detection instead of traditional bounding box regression.

\myparagraph{End-to-end object detection.} 
The well-established end-to-end object detectors are based on sparse candidates and multiple-stage refinement. Relation Network~\cite{RelationNetworks} and DETR~\cite{DETR} directly output the predictions without any hand-crafted assignment and post-processing procedure, achieving fantastic performance. DETR utilizes a sparse set of object queries to interact with the global image feature. Benefit from the global attention mechanism~\cite{vaswani2017attention} and the bipartite matching between predictions and ground truth objects, DETR can discard the NMS procedure while achieving remarkable performance. Deformable-DETR~\cite{deformabledetr} is introduced to restrict each object query to a small set of crucial sampling points around the reference points, instead of all points in the feature map. Sparse R-CNN~\cite{sun2020sparse} starts from a fixed sparse set of learned object proposals and iteratively performs classification and localization to the object recognition head. Adaptive Clustering Transformer~\cite{ACT} proposes to improve the attention in DETR’s encoder by LSH approximate clustering. UP-DETR~\cite{UPDETR} improves the convergence speed of DETR by a self-supervised method. TSP~\cite{TSPRCNN} analyzes co-attention and bipartite matching are two main causes of slow convergence in DETR. SMCA~\cite{SMCA} explores global information with a self-attention and co-attention mechanism to achieve fast convergence and better accuracy performance.

\section{Conclusion}

Assigning positive samples by location cost is conceptually intuitive and popularizes in object detection to date. However, in this work, we surprisingly find that this widely-used method is the obstacle of end-to-end detectors. By additionally considering classification cost, previous detectors immediately achieve end-to-end detection. Our findings uncover that answer to the notorious problem of defining positive samples in object detection is embarrassingly simple: in every training iteration, selecting only one positive sample which could minimize training loss is just 'right'.

\myparagraph{Acknowledgements} 

This work was supported by the General Research Fund of HK No.27208720.

\bibliography{onenet}

\begin{thebibliography}{47}
\providecommand{\natexlab}[1]{#1}
\providecommand{\url}[1]{\texttt{#1}}
\expandafter\ifx\csname urlstyle\endcsname\relax
  \providecommand{\doi}[1]{doi: #1}\else
  \providecommand{\doi}{doi: \begingroup \urlstyle{rm}\Url}\fi

\bibitem[Burton et~al.(1997)Burton, Herold~G, and
  Rienk~S]{burton1997perceptron}
Burton, R.~M., Herold~G, D., and Rienk~S, V.
\newblock Perceptron algorithms for the classification of non-separable
  populations.
\newblock \emph{Stochastic Models}, 13\penalty0 (2):\penalty0 205--222, 1997.

\bibitem[Cai \& Vasconcelos(2018)Cai and Vasconcelos]{CascadeRCNN}
Cai, Z. and Vasconcelos, N.
\newblock Cascade {R-CNN}: Delving into high quality object detection.
\newblock In \emph{CVPR}, 2018.

\bibitem[Carion et~al.(2020)Carion, Massa, Synnaeve, Usunier, Kirillov, and
  Zagoruyko]{DETR}
Carion, N., Massa, F., Synnaeve, G., Usunier, N., Kirillov, A., and Zagoruyko,
  S.
\newblock {End-to-End} object detection with transformers.
\newblock In \emph{ECCV}, 2020.

\bibitem[Dai et~al.(2016)Dai, Li, He, and Sun]{R-FCN}
Dai, J., Li, Y., He, K., and Sun, J.
\newblock {R-FCN}: Object detection via region-based fully convolutional
  networks.
\newblock In \emph{NeurIPS}, 2016.

\bibitem[Dai et~al.(2020)Dai, Cai, Lin, and Chen]{UPDETR}
Dai, Z., Cai, B., Lin, Y., and Chen, J.
\newblock Up-detr: Unsupervised pre-training for object detection with
  transformers.
\newblock \emph{arXiv preprint arXiv:2011.09094}, 2020.

\bibitem[Dalal \& Triggs(2005)Dalal and Triggs]{hog}
Dalal, N. and Triggs, B.
\newblock Histograms of oriented gradients for human detection.
\newblock In \emph{CVPR}, 2005.

\bibitem[Everingham et~al.(2010)Everingham, Van~Gool, Williams, Winn, and
  Zisserman]{PASCAL-VOC}
Everingham, M., Van~Gool, L., Williams, C. K.~I., Winn, J., and Zisserman, A.
\newblock The pascal visual object classes ({VOC}) challenge.
\newblock \emph{IJCV}, 88\penalty0 (2):\penalty0 303--338, 2010.

\bibitem[Felzenszwalb et~al.(2010)Felzenszwalb, Girshick, McAllester, and
  Ramanan]{dpm}
Felzenszwalb, P., Girshick, R., McAllester, D., and Ramanan, D.
\newblock Object detection with discriminatively trained part based models.
\newblock \emph{T-PAMI}, 32\penalty0 (9):\penalty0 1627--1645, 2010.

\bibitem[Gao et~al.(2021)Gao, Zheng, Wang, Dai, and Li]{SMCA}
Gao, P., Zheng, M., Wang, X., Dai, J., and Li, H.
\newblock Fast convergence of detr with spatially modulated co-attention.
\newblock \emph{arXiv preprint arXiv:2101.07448}, 2021.

\bibitem[Girshick(2015)]{FastRCNN}
Girshick, R.
\newblock Fast {R-CNN}.
\newblock In \emph{ICCV}, 2015.

\bibitem[Girshick et~al.(2014)Girshick, Donahue, Darrell, and Malik]{RCNN}
Girshick, R., Donahue, J., Darrell, T., and Malik, J.
\newblock Rich feature hierarchies for accurate object detection and semantic
  segmentation.
\newblock In \emph{CVPR}, 2014.

\bibitem[He et~al.(2016)He, Zhang, Ren, and Sun]{ResNet}
He, K., Zhang, X., Ren, S., and Sun, J.
\newblock Deep residual learning for image recognition.
\newblock In \emph{CVPR}, 2016.

\bibitem[He et~al.(2017)He, Gkioxari, Dollar, and Girshick]{MaskRCNN}
He, K., Gkioxari, G., Dollar, P., and Girshick, R.
\newblock Mask {R-CNN}.
\newblock In \emph{ICCV}, 2017.

\bibitem[Hu et~al.(2018)Hu, Gu, Zhang, Dai, and Wei]{RelationNetworks}
Hu, H., Gu, J., Zhang, Z., Dai, J., and Wei, Y.
\newblock Relation networks for object detection.
\newblock In \emph{CVPR}, 2018.

\bibitem[Huang et~al.(2017)Huang, Liu, Van Der~Maaten, and
  Weinberger]{DenseNet}
Huang, G., Liu, Z., Van Der~Maaten, L., and Weinberger, K.~Q.
\newblock Densely connected convolutional networks.
\newblock In \emph{CVPR}, 2017.

\bibitem[Huang et~al.(2015)Huang, Yang, Deng, and Yu]{DenseBox}
Huang, L., Yang, Y., Deng, Y., and Yu, Y.
\newblock {DenseBox}: Unifying landmark localization with end to end object
  detection.
\newblock \emph{arXiv preprint arXiv:1509.04874}, 2015.

\bibitem[Krizhevsky et~al.(2012)Krizhevsky, Sutskever, and Hinton]{AlexNet}
Krizhevsky, A., Sutskever, I., and Hinton, G.~E.
\newblock {ImageNet} classification with deep convolutional neural networks.
\newblock In \emph{NeurIPS}, 2012.

\bibitem[Kuhn(1955)]{kuhn1955hungarian}
Kuhn, H.~W.
\newblock {The Hungarian method for the assignment problem}.
\newblock \emph{NRL}, 2\penalty0 (1-2):\penalty0 83--97, 1955.

\bibitem[Law \& Deng(2018)Law and Deng]{CornerNet}
Law, H. and Deng, J.
\newblock {CornerNet}: Detecting objects as paired keypoints.
\newblock In \emph{ECCV}, 2018.

\bibitem[Lin et~al.(2014)Lin, Maire, Belongie, Hays, Perona, Ramanan, Dollár,
  and Zitnick]{COCO}
Lin, T.-Y., Maire, M., Belongie, S., Hays, J., Perona, P., Ramanan, D.,
  Dollár, P., and Zitnick, C.~L.
\newblock Microsoft {COCO}: Common objects in context.
\newblock In \emph{ECCV}, 2014.

\bibitem[Lin et~al.(2017{\natexlab{a}})Lin, Dollar, Girshick, He, Hariharan,
  and Belongie]{FPN}
Lin, T.-Y., Dollar, P., Girshick, R., He, K., Hariharan, B., and Belongie, S.
\newblock Feature pyramid networks for object detection.
\newblock In \emph{CVPR}, 2017{\natexlab{a}}.

\bibitem[Lin et~al.(2017{\natexlab{b}})Lin, Goyal, Girshick, He, and
  Dollar]{FocalLoss}
Lin, T.-Y., Goyal, P., Girshick, R., He, K., and Dollar, P.
\newblock Focal loss for dense object detection.
\newblock In \emph{ICCV}, 2017{\natexlab{b}}.

\bibitem[Liu et~al.(2016)Liu, Anguelov, Erhan, Szegedy, Reed, Fu, and
  Berg]{SSD}
Liu, W., Anguelov, D., Erhan, D., Szegedy, C., Reed, S., Fu, C.-Y., and Berg,
  A.~C.
\newblock {SSD}: Single shot multibox detector.
\newblock In \emph{ECCV}, 2016.

\bibitem[Lu et~al.(2019)Lu, Li, Yue, Li, and Yan]{GridRCNN}
Lu, X., Li, B., Yue, Y., Li, Q., and Yan, J.
\newblock Grid {R-CNN}.
\newblock In \emph{CVPR}, 2019.

\bibitem[Newell et~al.(2017)Newell, Huang, and Deng]{associative}
Newell, A., Huang, Z., and Deng, J.
\newblock Associative embedding: End-to-end learning for joint detection and
  grouping.
\newblock In \emph{Advances in neural information processing systems}, pp.\
  2277--2287, 2017.

\bibitem[Pang et~al.(2019)Pang, Chen, Shi, Feng, Ouyang, and Lin]{LibraRCNN}
Pang, J., Chen, K., Shi, J., Feng, H., Ouyang, W., and Lin, D.
\newblock Libra {R-CNN}: Towards balanced learning for object detection.
\newblock In \emph{CVPR}, 2019.

\bibitem[Redmon \& Farhadi(2017)Redmon and Farhadi]{YOLO9000}
Redmon, J. and Farhadi, A.
\newblock {YOLO9000}: Better, faster, stronger.
\newblock In \emph{CVPR}, 2017.

\bibitem[Redmon et~al.(2016)Redmon, Divvala, Girshick, and Farhadi]{YOLO}
Redmon, J., Divvala, S., Girshick, R., and Farhadi, A.
\newblock You only look once: Unified, real-time object detection.
\newblock In \emph{CVPR}, 2016.

\bibitem[Ren et~al.(2015)Ren, He, Girshick, and Sun]{FasterRCNN}
Ren, S., He, K., Girshick, R., and Sun, J.
\newblock Faster {R-CNN}: Towards real-time object detection with region
  proposal networks.
\newblock In \emph{NeurIPS}, 2015.

\bibitem[Rezatofighi et~al.(2019)Rezatofighi, Tsoi, Gwak, Sadeghian, Reid, and
  Savarese]{GIoU}
Rezatofighi, H., Tsoi, N., Gwak, J., Sadeghian, A., Reid, I., and Savarese, S.
\newblock Generalized intersection over union: A metric and a loss for bounding
  box regression.
\newblock In \emph{CVPR}, 2019.

\bibitem[Shao et~al.(2018)Shao, Zhao, Li, Xiao, Yu, Zhang, and Sun]{crowdhuman}
Shao, S., Zhao, Z., Li, B., Xiao, T., Yu, G., Zhang, X., and Sun, J.
\newblock Crowdhuman: A benchmark for detecting human in a crowd.
\newblock \emph{arXiv preprint arXiv:1805.00123}, 2018.

\bibitem[Simonyan \& Zisserman(2015)Simonyan and Zisserman]{vgg}
Simonyan, K. and Zisserman, A.
\newblock Very deep convolutional networks for large-scale image recognition.
\newblock In \emph{ICLR}, 2015.

\bibitem[Sun et~al.(2020{\natexlab{a}})Sun, Zhang, Jiang, Kong, Xu, Zhan,
  Tomizuka, Li, Yuan, Wang, et~al.]{sun2020sparse}
Sun, P., Zhang, R., Jiang, Y., Kong, T., Xu, C., Zhan, W., Tomizuka, M., Li,
  L., Yuan, Z., Wang, C., et~al.
\newblock Sparse r-cnn: End-to-end object detection with learnable proposals.
\newblock \emph{arXiv preprint arXiv:2011.12450}, 2020{\natexlab{a}}.

\bibitem[Sun et~al.(2020{\natexlab{b}})Sun, Cao, Yang, and Kitani]{TSPRCNN}
Sun, Z., Cao, S., Yang, Y., and Kitani, K.
\newblock Rethinking transformer-based set prediction for object detection.
\newblock \emph{arXiv preprint arXiv:2011.10881}, 2020{\natexlab{b}}.

\bibitem[Szegedy et~al.(2015)Szegedy, Liu, Jia, Sermanet, Reed, Anguelov,
  Erhan, Vanhoucke, and Rabinovich]{GoogLeNet}
Szegedy, C., Liu, W., Jia, Y., Sermanet, P., Reed, S.~E., Anguelov, D., Erhan,
  D., Vanhoucke, V., and Rabinovich, A.
\newblock Going deeper with convolutions.
\newblock In \emph{CVPR}, 2015.

\bibitem[Tian et~al.(2019)Tian, Shen, Chen, and He]{FCOS}
Tian, Z., Shen, C., Chen, H., and He, T.
\newblock {FCOS}: Fully convolutional one-stage object detection.
\newblock In \emph{ICCV}, 2019.

\bibitem[Uijlings et~al.(2013)Uijlings, Van De~Sande, Gevers, and
  Smeulders]{SelectiveSearch}
Uijlings, J.~R., Van De~Sande, K.~E., Gevers, T., and Smeulders, A.~W.
\newblock Selective search for object recognition.
\newblock \emph{IJCV}, 104\penalty0 (2):\penalty0 154--171, 2013.

\bibitem[Vaswani et~al.(2017)Vaswani, Shazeer, Parmar, Uszkoreit, Jones, Gomez,
  Kaiser, and Polosukhin]{vaswani2017attention}
Vaswani, A., Shazeer, N., Parmar, N., Uszkoreit, J., Jones, L., Gomez, A.~N.,
  Kaiser, {\L}., and Polosukhin, I.
\newblock Attention is all you need.
\newblock In \emph{Advances in neural information processing systems}, pp.\
  5998--6008, 2017.

\bibitem[Viola \& Jones(2001)Viola and Jones]{boosted}
Viola, P. and Jones, M.
\newblock Rapid object detection using a boosted cascade of simple features.
\newblock In \emph{Proceedings of the 2001 IEEE computer society conference on
  computer vision and pattern recognition. CVPR 2001}, volume~1, pp.\  I--I.
  IEEE, 2001.

\bibitem[Vu et~al.(2019)Vu, Jang, Pham, and Yoo]{CascadeRPN}
Vu, T., Jang, H., Pham, T.~X., and Yoo, C.~D.
\newblock Cascade {RPN}: Delving into high-quality region proposal network with
  adaptive convolution.
\newblock In \emph{NeurIPS}, 2019.

\bibitem[Wang et~al.(2020)Wang, Song, Li, Sun, Sun, and Zheng]{defcn}
Wang, J., Song, L., Li, Z., Sun, H., Sun, J., and Zheng, N.
\newblock End-to-end object detection with fully convolutional network.
\newblock \emph{arXiv preprint arXiv:2012.03544}, 2020.

\bibitem[Zhang et~al.(2020{\natexlab{a}})Zhang, Chang, Ma, Wang, and
  Chen]{DynamicRCNN}
Zhang, H., Chang, H., Ma, B., Wang, N., and Chen, X.
\newblock Dynamic {R-CNN}: Towards high quality object detection via dynamic
  training.
\newblock In \emph{ECCV}, 2020{\natexlab{a}}.

\bibitem[Zhang et~al.(2019)Zhang, Xiong, Sun, Hu, Li, and Yu]{doubleanchor}
Zhang, K., Xiong, F., Sun, P., Hu, L., Li, B., and Yu, G.
\newblock Double anchor r-cnn for human detection in a crowd.
\newblock \emph{arXiv preprint arXiv:1909.09998}, 2019.

\bibitem[Zhang et~al.(2020{\natexlab{b}})Zhang, Chi, Yao, Lei, and Li]{ATSS}
Zhang, S., Chi, C., Yao, Y., Lei, Z., and Li, S.~Z.
\newblock Bridging the gap between anchor-based and anchor-free detection via
  adaptive training sample selection.
\newblock In \emph{CVPR}, 2020{\natexlab{b}}.

\bibitem[Zheng et~al.(2020)Zheng, Gao, Wang, Li, and Dong]{ACT}
Zheng, M., Gao, P., Wang, X., Li, H., and Dong, H.
\newblock End-to-end object detection with adaptive clustering transformer.
\newblock \emph{arXiv preprint arXiv:2011.09315}, 2020.

\bibitem[Zhou et~al.(2019)Zhou, Wang, and Kr{\"{a}}henb{\"{u}}hl]{CenterNet}
Zhou, X., Wang, D., and Kr{\"{a}}henb{\"{u}}hl, P.
\newblock Objects as points.
\newblock \emph{arXiv preprint arXiv:1904.07850}, 2019.

\bibitem[Zhu et~al.(2020)Zhu, Su, Lu, Li, Wang, and Dai]{deformabledetr}
Zhu, X., Su, W., Lu, L., Li, B., Wang, X., and Dai, J.
\newblock Deformable detr: Deformable transformers for end-to-end object
  detection.
\newblock \emph{arXiv preprint arXiv:2010.04159}, 2020.

\end{thebibliography}
\bibliographystyle{icml2021}

\newpage

\appendix

\section{Proof for Theoretical Analysis}

We focus on analyzing properties using linear classifier. Let $\mathcal{X}=\{x\in\mathbb{R}^d : \left\|x\right\| \leq 1 \}$ be an instance space and $\mathcal{Y} = \{+1, -1\}$ be the label space. The label of a positive sample is $+1$ while that of a negative sample is $-1$. We wish to train a classifier $h$, coming from a hypothesis class $\mathcal{H}=\{x \mapsto \mathrm{sign}( w\tran x) : w\in\mathbb{R}^d\}$. Note that we can express the bias term $b$ by rewriting $w=[\hat{w},b]\tran$ and $x=[\hat{x}, 1]\tran$. We use the perceptron’s update rule with mini-batch size of 1. That is, given the classifier $w_t\in\mathbb{R}^d$, the
update is only performed on incorrectly classified example $(x_t,y_t)\in \mathcal{X} \times \mathcal{Y}$ as given by $w_{t+1} = w_t +\eta y_tx_t$ where $\eta$ is the stepsize. 

\vspace{5mm}
\textbf{Proposition 4.2 (Feasibility)}
\textit{Suppose that the one-to-one assignment is run on a sequence of examples from $\mathcal{X} \times \mathcal{Y}$. Given weight vector $w_t = [\hat{w}_{t}, b_{t}]\tran$ at update step $t$, there exists $\gamma_t\in \mathbb{R}$ and $\delta_t >0$ such that for all $(x,y) \in \mathcal{X} \times \mathcal{Y}$ we have $yw^*_t\tran x \geq \delta_t$ with $w_t^* = [\hat{w}_{t}, \gamma_t]\tran$.}

\textit{Proof.} we denote $x_t^1 =\arg\max_{x\in\mathcal{X}} w_t\tran x$  and $x_t^2 =\arg\max_{x\in\mathcal{X}\backslash \{x_t^1\}} w_t\tran x$. We assume $w_t^1\tran x >0$, hence we can infer that $w_t\tran x_t^1 > w_t\tran x_t^2 >0$, otherwise the algorithm converges at $w_t$ because it satisfies that  $w_t\tran x_t^1>0$ and $w_t\tran x\leq 0$ for all $x\in\mathcal{X}\backslash \{x_t^1\}$. 
By one-to-one assignment, the label of $x_t^1$ is $y(x_t^1)=+1$ and the labels of the remaining samples in $\mathcal{X}$ are $y(x)=-1, x\in\mathcal{X}\backslash \{x_t^1\}$. 
Take $\gamma_t=-\frac{\hat{w}_t\tran(\hat{x}_1+\hat{x}_2)}{2}$, we have 
\vspace{-0.1in}
\begin{equation}\label{eq:delta1}
    \begin{aligned}
        y(x_t^1)w^*_t\tran x_t^1 &=\hat{w}_t\tran\hat{x}_t^1 -\frac{\hat{w}_t\tran(\hat{x}_1+\hat{x}_2)}{2}\\
        &= \frac{\hat{w}_t\tran(\hat{x}_1-\hat{x}_2)}{2} > 0
    \end{aligned}
\end{equation}
\vspace{-0.1in}
and for all $x\in\mathcal{X}\backslash \{x_t^1\}$ we have
\begin{equation}\label{eq:delta2}
    \begin{aligned}
        y(x)w^*_t\tran x &=-1* (\hat{w}_t\tran\hat{x} -\frac{\hat{w}_t\tran(\hat{x}_1+\hat{x}_2)}{2})\\
        &\geq \frac{\hat{w}_t\tran(\hat{x}_1-\hat{x}_2)}{2} > 0
    \end{aligned}
\end{equation}
where the first inequality holds by $w_t\tran x \leq w_t\tran x_t^2$ since $x_t^2 =\arg\max_{x\in\mathcal{X}\backslash \{x_t^1\}} w_t\tran x$.

By Eqn.(\ref{eq:delta1}) and Eqn.(\ref{eq:delta2}), we can take $\delta_t=\frac{\hat{w}_t\tran(\hat{x}_1-\hat{x}_2)}{2}$.

\vspace{5mm}
\textbf{Theorem 4.3 (Convergence)}
\textit{ Let $\gamma_{t+1}$ and $\gamma_t$ be the constants defined in Proposition 4.2. For each update step $t$, we assume there exists a stepsize $\eta_t$ such that $\left\|x_t\right\|^2\eta_t^2+y_t(\gamma_{t+1}-2\gamma_{t})\eta_t+b_t(\gamma_{t+1}-\gamma_{t})>0$ where $(x_t, y_t)$ be the incorrectly classified sample at iteration $t$. 
If the sample label is assigned by one-to-one assignment, then, $t \leq \frac{{\eta_{max}^2}-2\eta_{min}\delta_{\min}(w_{1}\tran w_{0}^*-\left\|w_{0}\right\| -{\eta_{max}})}{2\eta_{min}^2\delta_{\min}^2}$ where $\eta_{max}$ and $\eta_{min}$ are the maximum and minimum value of stepsize among all $t$'s updates, $w_1$  is the classifier after the first update and $\delta_{\min}$ is the minimum of all $\delta_t$s in Proposition 4.2. All instances at initialization can be correctly classified by $w_0^*$.
} 

We first show  $w_{t+1}\tran w_{t+1}^* \geq w_{t+1}\tran w_{t}^*$.  Rewriting the weight vector $w_t$ into a normal vector and a bias gives us
\begin{equation}\label{eq:normal-bias}
    \left[\begin{matrix} \hat{w}_{t+1} \\ b_{t+1} \end{matrix}\right] =  \left[\begin{matrix} \hat{w}_{t} \\ b_{t} \end{matrix}\right] +  \eta y_t\left[\begin{matrix} \hat{x}_{t} \\ 1 \end{matrix}\right]
\end{equation}
From Eqn.(\ref{eq:normal-bias}), we have $w_{t+1}=[\hat{w}_t+\eta y_t\hat{x}_t, b_t+\eta y_t]\tran$ at update $t$. According to the definition of $\gamma_t$ and $\gamma_{t+1}$, we obtain $w_{t+1}^*=[\hat{w}_t+y_t\hat{x}_t, \gamma_{t+1}]$ and $w_t^*=[\hat{w}_t,\gamma_{t}]$. Therefore, we can derive that
\begin{equation}\label{eq:wtdeduction}
\begin{aligned}
    &w_{t+1}\tran w_{t+1}^* - w_{t+1}\tran w_{t}^* \\ 
    &=(\hat{w}_t+\eta y_t\hat{x}_t)\tran \eta  y_t\hat{x}_t
    +(b_t+\eta y_t)(\gamma_{t+1}-\gamma_t) \\
    &=\left\|x_t \right\|^2\eta^2 +y_t(\hat{w}_t\hat{x}_t-\gamma_{t}+\gamma_{t+1})\eta +b_t(\gamma_{t+1}-\gamma_{t})\\
    &\geq \left\|x_t \right\|^2\eta^2 +y_t(\gamma_{t+1} - 2\gamma_{t})\eta +b_t(\gamma_{t+1}-\gamma_{t})
\end{aligned}
\end{equation}

Taking $\eta=\eta_t$ gives us $w_{t+1}\tran w_{t+1}^* \geq w_{t+1}\tran w_{t}^*$ by the assumption. Note that the assumption in 
Theorem 4.3 easily holds when $\eta_t$ is a large but finite number due to the property of quadratic equation of one variable in Eqn.(\ref{eq:wtdeduction}).

To proceed, we find upper and lower
bounds on the length of the weight vector $w_t$ to show
finite number of updates.
By convenience, we normalize $w_{t}^*$ to $\left\|w_{t}^*\right\|$ = 1.
Assume that after $t+1$ steps the weight vector $w_{t+1}$ has been computed.
This means that at time $t$ a training sample was incorrectly classified by the weight vector $w_t$ and so $w_{t+1} = w_t + \eta_ty_tx_t$. By one-to-one assignment, we have $y_t = 1$ if $x_t = \arg\max_{x\in\mathcal{X}} w_t\tran x$ and $-1$ otherwise.

By computing the length of $w_{t+1}$, we arrive at
\begin{equation}
\begin{aligned}
    \left\|w_{t+1}\right\|^2 &= (w_t + \eta_t y_tx_t)\tran (w_t + \eta_t y_tx_t)\\
    &= \left\|w_{t}\right\|^2 + \left\|x_{t}\right\|^2\eta_t^2 + 2y_t w_t\tran x_t\eta_t \\
    & \leq \left\|w_{t}\right\|^2 + \eta_t^2
\end{aligned}\label{eq:upper1}
\end{equation}
where the third equation holds because the length of instance $x$ is bounded by $1$ and $y_t w_t\tran x_t$ is negative or zero (otherwise we would have not corrected $w_t$
using sample $(x_t,y_t)$ by perceptron's update rule) . Induction through Eqn.(\ref{eq:upper1}) then gives us
\begin{equation}\label{eq:up-bound}
    \left\|w_{t+1}\right\|^2 \leq \left\|w_{0}\right\|^2 + \sum_{k=0}^t\eta_k^2\leq (t+1)\eta_{max}^2
\end{equation}
where $\eta_{min}=\max\{\eta_k: k=0,1,\cdots, t\}$.
To drive the lower bound,  we multiply $w_t^*$ in Proposition 4.2 on both sides of $w_{t+1} = w_t + \eta_t y_tx_t$,
it gives us $w_{t+1}\tran w_{t}^* = w_{t}\tran w_{t}^* + \eta_t y_t w_t^*\tran x_t$. By Eqn.(\ref{eq:wtdeduction}), it can be relaxed into
\begin{equation}\label{eq:low-bound2}
    \begin{aligned}
        w_{t+1}\tran w_{t}^* &= w_{t}\tran( w_{t}^*-w_{t-1}^*+w_{t-1}^*) +\eta_ty_t w_t^*\tran x_t\\
        &= w_{t}\tran w_{t-1}^*+w_{t}\tran( w_{t}^*-w_{t-1}^*)+\eta_ty_t w_t^*\tran x_t\\
        &\geq w_{t}\tran w_{t-1}^*+\eta_ty_t w_t^*\tran x_t\\
        &\geq w_{t}\tran w_{t-1}^*+\eta_t\delta_t
    \end{aligned}
\end{equation}
where the first inequality holds by Eqn.(\ref{eq:wtdeduction}), the second inequality holds by Proposition 4.2. Induction through Eqn.(\ref{eq:low-bound2}) then yields
\begin{equation}\label{eq:low-bound}
    w_{t+1}\tran w_{t}^* \geq w_{1}\tran w_{0}^* + \sum_{k=1}^t\eta_k\delta_k \geq w_{1}\tran w_{0}^* +t\eta_{min}\delta_{min}
\end{equation}
where $\delta_{min}=\min\{\delta_k: k=1,\cdots, t\}$ and $\eta_{min}=\min\{\eta_k: k=1,\cdots, t\}$. Combining Eqn.(\ref{eq:up-bound}) and Eqn.(\ref{eq:low-bound}), we obtain that
\begin{equation}
    w_{1}\tran w_{0}^*+t\eta_{min}\delta_{\min} \leq \sqrt{\left\|w_{0}\right\|^2 + (t+1)\eta_{max}^2}
\end{equation}
Using $\sqrt{a+b}\leq\sqrt{a}+\sqrt{b}$, the above implies that
\begin{equation}
    w_{1}\tran w_{0}^* + t\eta_{min}\delta_{\min} \leq \left\|w_{0}\right\| + \sqrt{t}\eta_{max}+{\eta_{max}}
\end{equation}
Using standard algebraic manipulations, the above implies that
\begin{equation}
\begin{aligned}
     t &\leq (\frac{{\eta_{max}}+\sqrt{\eta_{max}^2-4\eta_{min}\delta_{\min}(w_{1}\tran w_{0}^*-\left\|w_{0}\right\| -{\eta_{max}})}}{2\eta_{min}\delta_{\min}})^2\\
     &\leq \frac{{\eta_{max}^2}-2\eta_{min}\delta_{\min}(w_{1}\tran w_{0}^*-\left\|w_{0}\right\| -{\eta_{max}})}{2\eta_{min}^2\delta_{\min}^2}
\end{aligned}
\end{equation}
This completes the proof.

\section{Positive Samples for Multiple Objects}

\begin{figure}[h!]
\centering
     \includegraphics[width=0.43\textwidth]{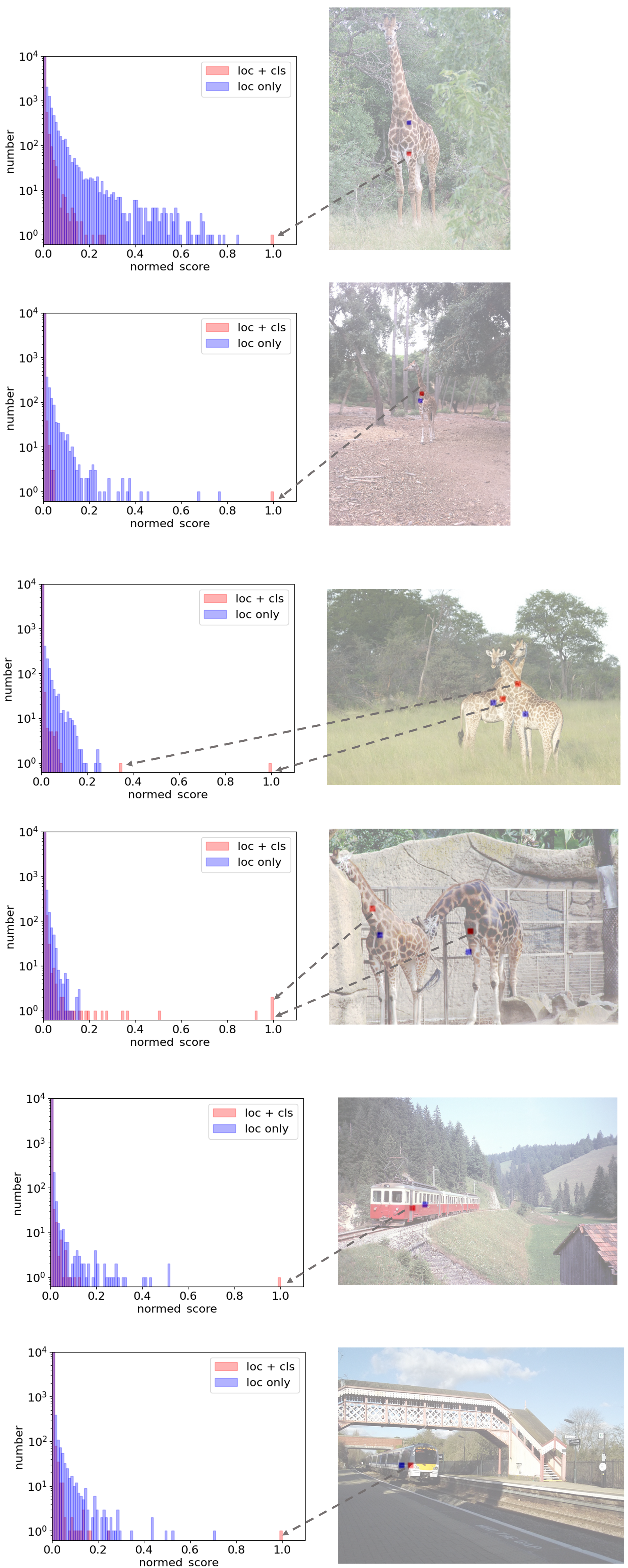}
     \vspace{2mm}
     \caption{\textbf{Positive samples in different training images.} For better visualization, we only show the part below the number of $10^4$, and scores are normalized to [0, 1]. Blue bins show the detector trained with positive samples chosen by only location cost. Red bins consider both location cost and classification cost. For multiple objects, classification cost produces two clusters of samples, the scores of positive sample cluster are obviously higher than samples in negative sample cluster.}
\label{fig:appendix_pos}
\end{figure}

As discussed in Section 4, when there exists an object in the image, classification cost results in a clear score gap between the sample of the first-highest score and the sample of the second-highest score. In Figure~\ref{fig:appendix_pos}, we show positive sample for multiple objects. Classification cost produces two clusters of samples, one of which is composed of positive samples, and their scores are obviously higher than samples in another cluster.

\begin{figure*}[!t]
\centering
\begin{subfigure}{1.00\textwidth}
    \centering
    \includegraphics[width=1.00\textwidth]{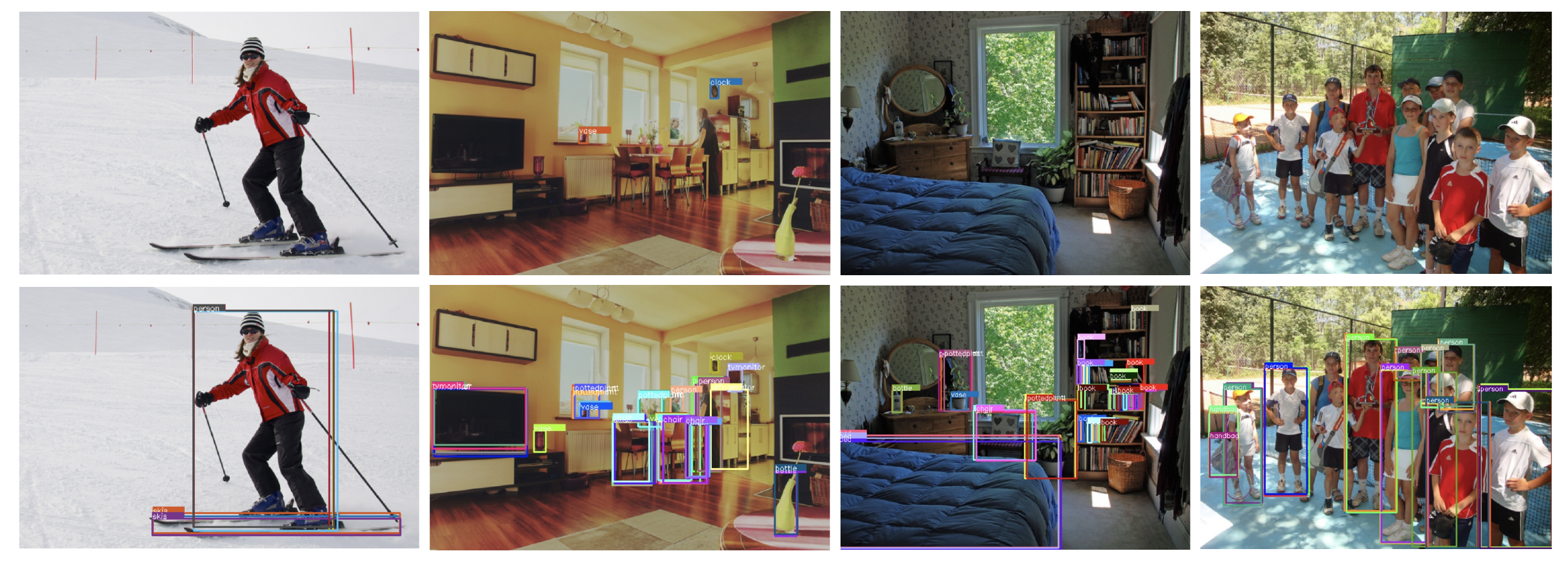}
    \caption{\textbf{Cost is location cost only}. 1st row shows boxes whose classification scores are higher than 0.4, 2nd shows 0.2.}
    \label{fig:app_1a}	
\end{subfigure} 
\vspace{2mm}

\begin{subfigure}{1.00\textwidth}
     \centering
     \includegraphics[width=1.00\textwidth]{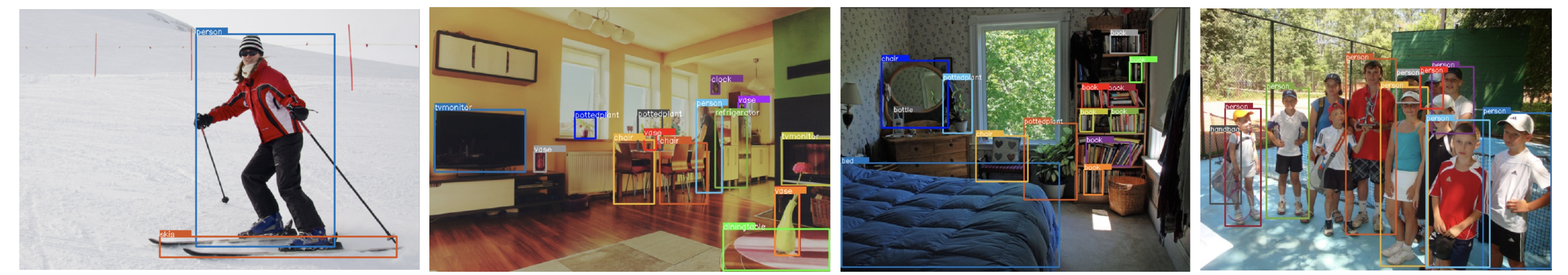}
     \caption{\textbf{Cost is summation of location cost and classification cost}. Those boxes whose classification scores are higher than 0.4 are shown. }     
     \label{fig:app_1b}
\end{subfigure}
\caption{\textbf{The advantage of large score gap to end-to-end object detection.} In figure (a), non-maximum predictions cannot be easily filtered out: a high threshold may filter out all predictions, while a low threshold may output multiple predictions. In figure (b), non-maximum predictions can be easily filtered out by a well-chosen score threshold.} 
\label{fig:app_score_gap}
\end{figure*}

\begin{table}[t]
\begin{center}
\setlength{\tabcolsep}{0.8mm}
\begin{tabular}
{ l | c  | c | c c c }
\toprule
Method &
Matching & 
COCO AP &
AP$_{50}$   & 
mMR$\downarrow$  & 
Recall \\
\midrule
RetinaNet$\dag$ & Bipartite & 37.5 & 90.8 & 49.3 & 98.1 \\

& MinCost & 37.5 & 81.3 & 76.7 & 95.1\\

FCOS$\dag$ & Bipartite & 38.9 & 90.7 & 48.2 & 97.6 \\

& MinCost &  38.9 & 80.5 & 79.3 & 94.1 \\
\bottomrule
\end{tabular}
\end{center}
\vspace{-4mm}
\caption{\textbf{Comparisons of bipartite matching and MinCost Matching.} $\dag$ is with one-to-one positive sample assignment and classification cost. MinCost works well when sample density is largely more than ground-truth object on COCO dataset. However, MinCost doesn't work on CrowdHuman, where multiple ground-truth objects may share the same positive sample. }
\label{table:mincost}
\vspace{-2mm}
\end{table}

\section{Large Score Gap}
We illustrate visualization examples in Figure~\ref{fig:app_score_gap} to show the advantage of large score gap to end-to-end object detection. In Figure~\ref{fig:app_score_gap}(a), the matching cost is only location cost, and the score gap is negligible, making non-maximum predictions cannot be easily filtered out: a high threshold(0.4), may filter out all predictions, while a low threshold(0.2), outputs multiple predictions. Instead, in Figure~\ref{fig:app_score_gap}(b),  additionally considering classification cost produces a large score gap, then non-maximum predictions can be filtered out by a well-chosen score threshold(0.4), achieving a successful end-to-end object detection.

\section{Bipartite Matching and MinCost Matching}
End-to-end object detection requires one-to-one positive sample assignment and bipartite-matching is usually used to prevent one sample assigned to many ground-truths. However, when density of object candidates are largely more than ground-truth objects in COCO dataset, simply selecting the one sample of minimum cost among all sample as the positive sample for each ground-truth object, termed as MinCost, is qualified to achieve the same performance as bipartite-matching, \eg, RetinaNet and FCOS. Nevertheless, MinCost doesn't work in CrowdHuman dataset, where samples density may not be greater than ground-truth objects, and multiple ground-truth objects may share the same positive sample. Therefore, bipartite-matching is still needed in crowded scenes.

\begin{figure*}[t]
\centering
\includegraphics[width=0.90\textwidth]{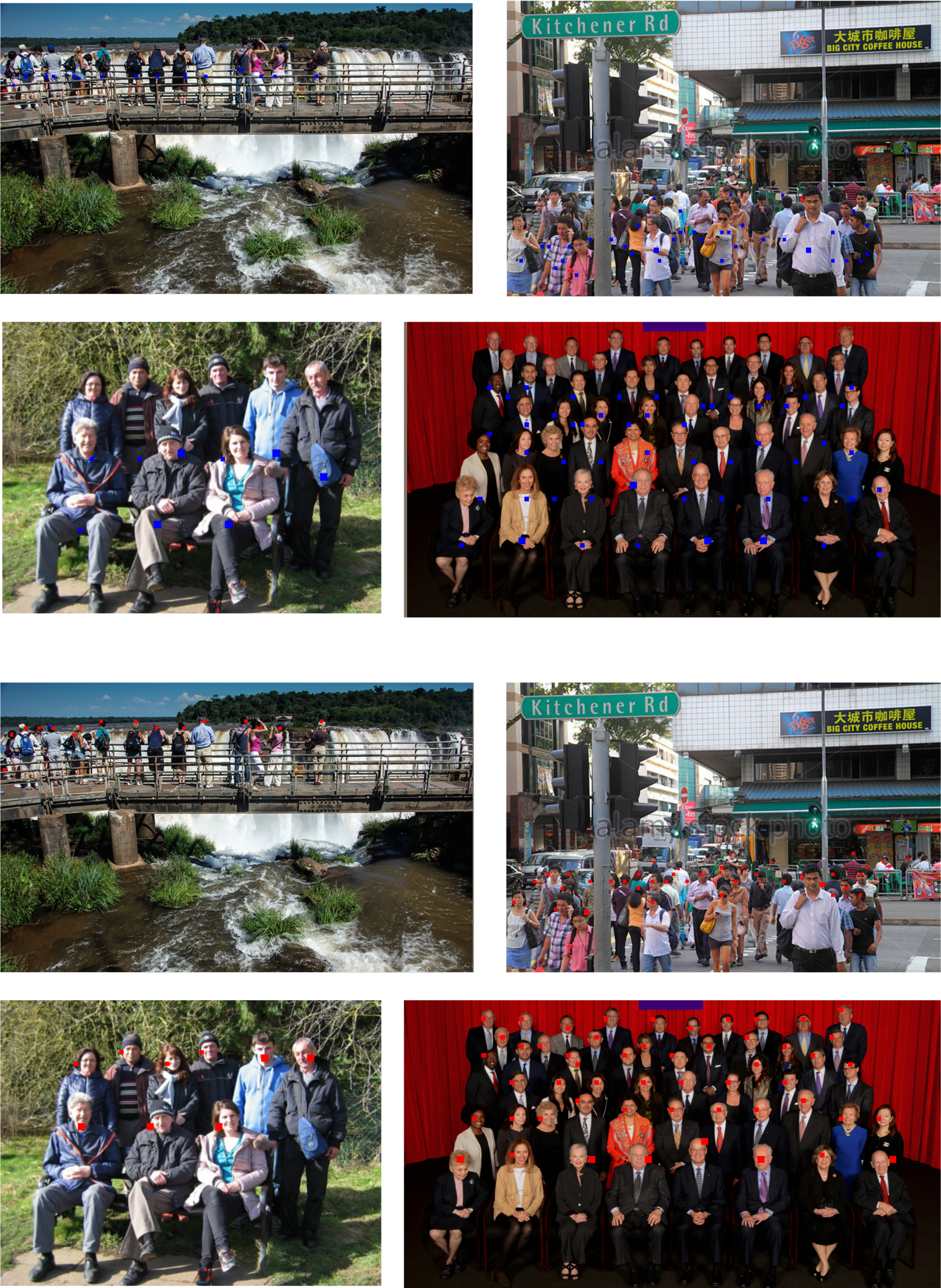}
\caption{\textbf{Positive samples in CrowdHuman images} by using FCOS~\cite{FCOS}, where the samples are the grid points in multi-layer feature maps. The blue points are chosen by only location cost, while the red points additionally consider classification cost. We surprisingly find that the red points are always the grid points in the area of human heads. This phenomenon is consistent with Figure~\ref{fig:pos}, where classification cost selects the positive samples in more discriminative area of the object.
}
\label{fig:app_crowdhuman}
\vspace{2mm}
\end{figure*}

\section{Weakly-Supervised Learning for Locating Human Head}

As discussed in Figure~\ref{fig:pos}, classification cost selects the positive samples in more discriminative area of the object. In CrowdHuman dataset, we surprisingly find that the chosen positive samples are always in the area of human heads when the samples are the grid points in multi-layer feature maps, \eg, FCOS, as shown in Figure~\ref{fig:app_crowdhuman}. This is because the head is the most discriminative part of the human. This finding may inspire new methods of weakly-unsupervised learning for locating human heads.

\end{document}